\documentclass[times,preprint,12pt]{elsarticle}

\usepackage{amsmath,amsfonts}
\usepackage{amssymb}
\usepackage{algorithmic}
\usepackage{algorithm}
\usepackage{array}
\usepackage[caption=false,font=normalsize,labelfont=sf,textfont=sf]{subfig}
\usepackage{textcomp}
\usepackage{url}
\usepackage{verbatim}
\usepackage{graphicx}
\usepackage{multirow}
\usepackage{booktabs}
\usepackage{ulem}
\usepackage{bbm}
\usepackage{csquotes}
\usepackage{cases}
\usepackage[left=3.2cm, right=3.2cm, top=3.8cm, bottom=3.8cm]{geometry}

\newtheorem{theorem}{Theorem}
\newtheorem{lemma}{Lemma}
\DeclareMathOperator{\diag}{diag}

\newenvironment{pf}[1][Proof]{\noindent {\bf #1.} }{\  \rule{0.5em}{0.5em}}

\begin{document}

\begin{frontmatter}

\title{Cross-Modality Learning on Heavily Noisy Attributed Graphs}

\author[xjtu]{Xi Wang}
\author[xjtu]{Limin Li\corref{cor1}}
\ead{liminli@mail.xjtu.edu.cn}
\author[sz]{Zhenyue Zhang}
\ead{zyzhang@zju.edu.cn}

\cortext[cor1]{Corresponding author.}

\address[xjtu]{School of Mathematics and Statistics, Xi'an Jiaotong University,
Xi'an, Shaanxi 710049, China}
\address[sz]{Faculty of Computational Mathematics and Cybernetics,
Shenzhen MSU-BIT University, Shenzhen 518172, China}

\begin{abstract}
Graph learning aims to recover latent topological structures from noisy graphs and auxiliary data. Existing attributed graph models typically learn hidden representations by integrating the node connections in graphs with the node attributes in auxiliary data, but their effectiveness is limited by heavy noise, which obscures genuine node relationships and hinders the discovery of intrinsic structures.
In this work, a novel framework of Cross-Modality Learning (CML) is proposed, motivated by our previous work on graph refinement. CML adopts a bidirectional cross-modality strategy to mutually exploit the complementary
information in noisy graphs and node attributes and recover the latent class information. Two mathematical models are developed within the CML framework: Attribute-driven CML (A-CML), which prioritizes node attributes and treats graph topology as auxiliary information, and Topology-driven CML (T-CML), which assumes that graph topology is more reliable than node attributes. Both models can be efficiently solved using alternating optimization algorithms with guaranteed convergence.
CML serves three roles: an attributed graph clustering method when combined with spectral projection, a cross-modality approach to single-view subspace learning, and a preprocessing method for other graph-based learning algorithms.
Extensive experiments, including simulation studies under asymmetric, high-noise settings and evaluations on six real-world attributed graph benchmarks against 15 state-of-the-art methods, demonstrate the recovery capability of the proposed CML framework.
\end{abstract}



\end{frontmatter}
\thispagestyle{plain}

\section{Introduction}

Attributed graphs, which integrate both the topological connections and multi-dimensional attributes of nodes, offer a comprehensive representation of complex real-world data and underpin downstream tasks such as node classification, link prediction, and graph classification. Typically, the primary objective in unsupervised graph learning is to recover the latent clustering structures from these graphs. Existing paradigms, ranging from graph neural networks (GNNs)~\cite{GNN}--\cite{GGs-NN} to non-parametric feature propagation methods like AGC~\cite{AGC}, heavily rely on integrating both structural connections and attribute features to learn node representations.

Based on how the two sources are integrated, existing unsupervised graph clustering methods can be categorized into two main streams. The first utilizes deep architectures for non-linear representation learning: GNNs propagate node attributes through stacked layers over topological neighborhoods, with GCN~\cite{GCN} acting as a first-order spectral convolution approximation and GAT~\cite{GAT} weighting neighbours by adaptive attention. Despite excelling at capturing complex regularities, these deep methods often suffer from structural uninterpretability, high parametric computational overhead, and deep-layer over-smoothing.

To alleviate the heavy reliance on network parameters and complex training phases, the second stream diverges from deep learning, focusing on parameter-free graph filtering and low-dimensional subspace recovery. A pioneering representative is AGC~\cite{AGC}, which utilizes the graph Laplacian matrix to directly smooth node features, offering an elegant, unsupervised, and parameter-free clustering methodology. However, directly employing the raw observed adjacency matrix $A$ for feature smoothing is highly vulnerable to topological distortions. To improve robustness, recent methods such as GCC~\cite{GCC}, FGC~\cite{FGC}, and SAGSC~\cite{SAGSC} shift their focus toward attribute matrices approximately sampled from a low-dimensional subspace, augmenting the raw feature propagation process with traditional subspace clustering techniques and enforcing low-rank or sparse regularization on the self-expressive matrix.

Since graphs in practice are rarely clean, robustness has become an active direction in its own right; recent designs, for instance, resist unreliable features and edges by coupling collaborative augmentation with contrastive learning~\cite{zhao2026hybrid}. In these pipelines the two modalities are still consumed jointly, and the asymmetry between their reliabilities is not exploited.

The problem addressed by these approaches is, however, distinct from that of multi-view subspace clustering (MVSC)~\cite{ARLRR, ESTMC, chen2026smhgc}, which seeks a consensus self-representation across multiple homogeneous feature views $\{X^{(v)}\}$ of the same samples. An attributed graph instead couples a single attribute mapping $X \in \mathbb{R}^{d \times n}$ with a pairwise adjacency matrix $A \in \mathbb{R}^{n \times n}$; since $A$ is indexed by sample pairs rather than by features, it is not a view in the MVSC sense and cannot be stacked with $X$. The two modalities are moreover corrupted asymmetrically and to different degrees, so that the restoration of one from the other must be directed rather than a symmetric fusion of the two sources.

\begin{figure}[t!]
\centering
\begin{minipage}{0.15\linewidth}
    \centering
    \includegraphics[width=\linewidth]{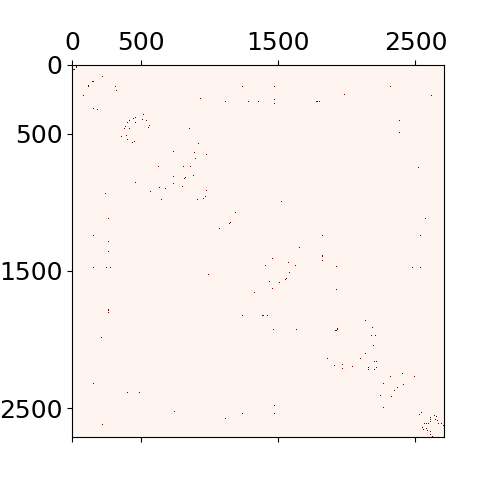}
    \\  \scriptsize (a) Sparse \(A\) 
\end{minipage}\hfill
\begin{minipage}{0.15\linewidth}
    \centering
    \includegraphics[width=\linewidth]{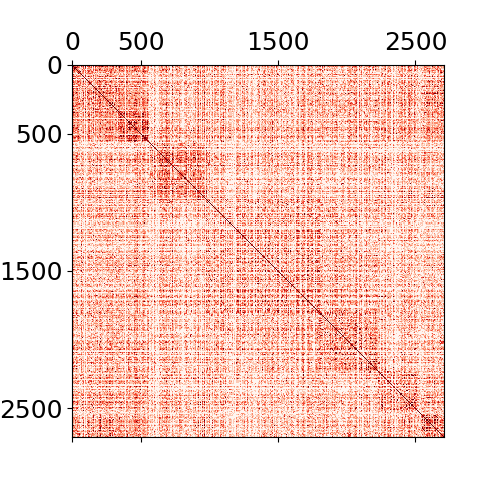}
    \\  \scriptsize (b) Attribute (\(X^\top X\))
\end{minipage}\hfill
\begin{minipage}{0.15\linewidth}
    \centering
    \includegraphics[width=\linewidth]{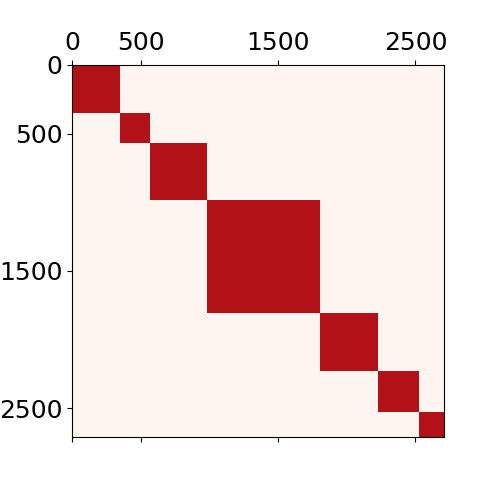}
    \\  \scriptsize (c) Ground truth
\end{minipage}\hfill
\begin{minipage}{0.15\linewidth}
    \centering
    \includegraphics[width=\linewidth]{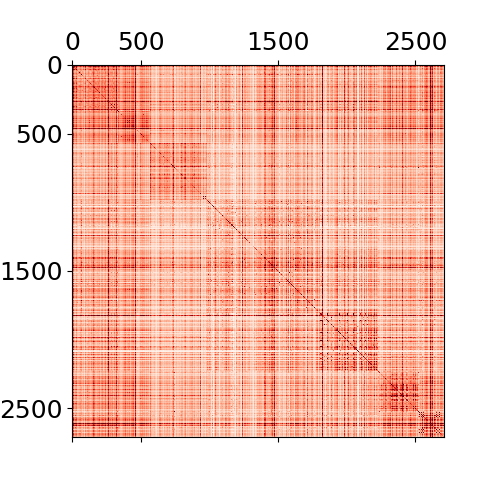}
    \\  \scriptsize (d) AGC
\end{minipage}\hfill
\begin{minipage}{0.15\linewidth}
    \centering
    \includegraphics[width=\linewidth]{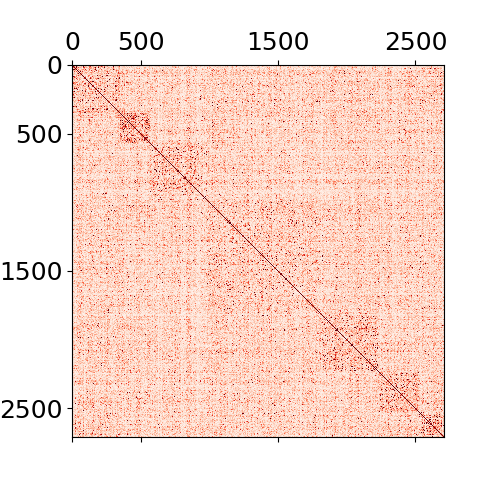}
    \\  \scriptsize (e) FGC
\end{minipage}\hfill
\begin{minipage}{0.15\linewidth}
    \centering
    \includegraphics[width=\linewidth]{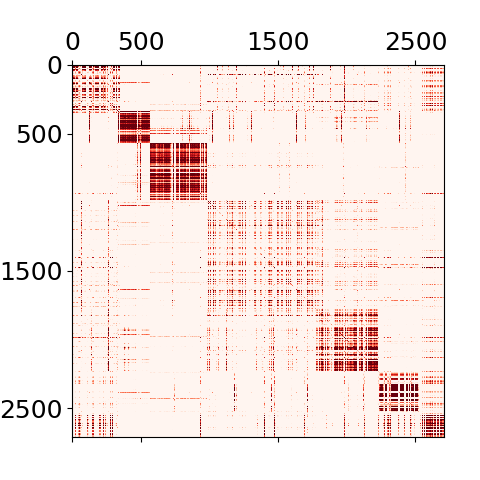}
    \\   \scriptsize (f) CML(Ours)
\end{minipage}

\caption{An illustrative example of CML challenges in a heavily noisy attributed graph.}
\label{fig:1}
\end{figure}

Both GNNs and the subspace-based graph filtering methods aim to find hidden graph structures by combining node features and connections. However, real-world data quality is often uneven, meaning either graph edges or node features are heavily corrupted. Under these conditions, standard methods fail because they rely on symmetric integration strategies and such blind blending allows severe noise from the damaged side to pollute the clean one. Consequently, the resulting node embeddings lose clear graphical interpretability and fail to capture true data relationships. To address this, the core challenge hinges on decoupling the true topology from heavy noise. 

A motivating example of this challenge and our solution is illustrated in Figure \ref{fig:1}. For an attributed graph $(A, X)$ generated from a ground-truth cluster structure, Figure \ref{fig:1}(a, b) plot the heavily noisy adjacency matrix and feature similarity matrix, respectively, while Figure \ref{fig:1}(c) presents the ideal block-diagonal structure. Standard embedding-based methods suffer from mutual noise contamination and fail to maintain structural clarity. For instance, the node similarity matrices derived from AGC (Figure \ref{fig:1}(d)) and FGC (Figure \ref{fig:1}(e)) exhibit highly blurred group structures. In contrast, the target graph $Z$ learned by our proposed method (Figure \ref{fig:1}(f)) successfully prevents mutual interference between modalities, recovering a clean, block-diagonal pattern that offers a highly interpretable and high-fidelity representation of the underlying clusters.

In this work, we present a framework of Cross-Modality Learning (CML), a novel dual-model 
approach designed for unsupervised graph learning under heavy noise. The CML framework achieves cross-modal mutual restoration through two complementary sub-models: Attribute-driven CML (A-CML) and Topology-driven CML (T-CML). Specifically, when topology is heavily perturbed, A-CML imposes sparse low-rank constraints to recover the true clustering alignment by exploiting the uncorrupted subspace structures of node attributes. Concurrently, when attributes suffer from severe feature-level noise, T-CML utilizes the relatively clean topological connections to cross-modally reconstruct the pristine attribute-space clustering. Both models resolve their resulting nonconvex optimization problems via an alternating iterative scheme equipped with proximal truncation operators, backed by a mathematically guaranteed convergence analysis. 
Extensive experiments, including simulation studies under asymmetrical noise configurations, evaluations on six benchmark graph datasets against 15 state-of-the-art methods, and extensions to single-input task, demonstrate that our framework consistently recovers stable clustering structures under diverse data corruptions.


\section{Related Work}
Attributed graph clustering considers both node features and structural information in order to comprehend complex graph data, and the landscape of existing solutions has been surveyed in~\cite{li2025clustering}. 

The emergence of GNNs led to various approaches that utilize graph encoders to produce new graph embeddings. To address the issue of unsupervised learning of graphs, Kipf and Welling~\cite{VGAE} introduce the Graph Autoencoder (GAE) together with its variational counterpart (VGAE), and Wang \textit{et al.}~\cite{MGAE} devise the Marginalized Graph Autoencoder (MGAE) for graph clustering via marginalized graph convolutional networks, while approaches such as ARGA, ARVGA~\cite{ARGA} and AGAE~\cite{AGAE} employ adversarial training mechanisms to enhance the discriminative power of sample representations. However, these methods are limited to capturing only the two or three hops away neighbours of each node, which may result in overlooking the global cluster structure in large graphs. DAEGC~\cite{DAEGC} counters this through deep attention mechanisms, while Simplified Graph Convolution (SGC)~\cite{SGC} and Simple Spectral Graph Convolution (S$^2$GC)~\cite{SSGC} reduce propagation to fixed low-pass filters and to a trade-off between low-pass and high-pass bands, respectively. A significant drawback is their dependence on pre-training with a clustering-guided loss, so that performance becomes contingent on the initial centers, and stacked layers encounter over-smoothing. SDCN~\cite{SDCN} and DFCN~\cite{DFCN} mitigate the second by learning both attribute and structure information, using information transport operations and structure-attribute fusion modules, and adaptive graph attention whose coefficients are optimized jointly with the clustering objective~\cite{duan2025adaptive} has been proposed for the same purpose; since the attention is estimated from the observed adjacency, the resulting fusion still inherits its noise.

Methods that need no training smooth the attributes directly: AGC~\cite{AGC} applies a low-pass filter of the observed adjacency, and FGC~\cite{FGC} shows that such propagation is simple yet effective. Smoothing with the raw $A$ is vulnerable to topological distortion, so GCC~\cite{GCC}, which aligns node representations with reconstructed cluster prototypes, and SAGSC~\cite{SAGSC}, which improves efficiency while jointly learning representations and assignments, place the self-expressive matrix on a low-dimensional subspace estimate with low-rank or sparse regularization; this formulation is now referred to as subspace clustering on attributed graphs (SCAG)~\cite{lin2025spectral}. Non-deep alternatives pursue efficiency through entropy-regularized matrix factorization~\cite{mohammadi2025dual} or anchor-based constrained Laplacian rank for large graphs~\cite{ma2025large}. What this line leaves implicit is the graph itself: the self-expressive matrix is
induced by a filtered representation and never constrained directly.
TGL~\cite{zheng2025one} instead couples a neighbour graph with a self-expressive
graph under a low-rank tensor constraint, but the constraint only asks the two graphs to agree; agreement is weaker than recovery, since it does not decide which source to trust when they disagree.

Recent methods incorporate the clustering objective into representation learning, generally assuming that node attributes and graph topology are jointly informative. Graph InfoClust (GIC)~\cite{GIC} maximizes intra-cluster mutual information, and newer designs boost weakly relevant features while refining the target distribution~\cite{liu2026adaptive}, or regularize the observed graph through structural entropy so that edges unrelated to clustering are down-weighted~\cite{zhang2025unsupervised}. These mechanisms act on the representation, on the partition, or on a reweighting of the observed adjacency; being designed for relatively clean observations, they often deteriorate when both topology and attributes are severely corrupted.

Taken together, the three lines share one premise: that the two modalities are equally trustworthy, that the graph is a by-product of a filtered representation, and that the two sources are fused directly rather than through a recovered graph. None of them therefore recovers the latent clustering structure hidden beneath heavy perturbations of either modality, and most overlook the joint modeling of sparsity, low-rankness, and non-negativity that interpretable structures require under noise. In contrast, CML makes the graph itself the object of recovery: the relatively clean modality drives the reconstruction of the corrupted one, and the framework returns both an embedding matrix for spectral clustering and an interpretable weighted graph $Z$ constrained to be sparse, low-rank and non-negative.

\section{Method}
\subsection{Problem Statement}
Given an undirected attributed graph $\mathcal{G}=\{A,X\}$ with $n$ nodes, the graph is represented by an empirical binary adjacency matrix $A=\{a_{ij}\}\in\mathbb{R}^{n\times n}$ and a high-dimensional node attribute matrix $X\in\mathbb{R}^{d\times n}$.
We assume the node representations intrinsically lie on a union of $K$ distinct low-dimensional subspaces, whose latent cluster structure is severely corrupted by structural anomalies in $A$. 

In real-world applications, data quality is often heavily noisy between features and graphs. Therefore, we study this task under two different bad scenarios. In the first scenario (graph-corrupted), the original graph $A$ is very noisy or breaks down, but the node features $X$ are relatively clean and reliable. In the second scenario (feature-corrupted), the features $X$ contain heavy noise and errors, but a relatively high-quality graph $A$ is well-preserved. How to solve both conflicting problems within a single, unified model is the main challenge of this work.

\subsection{
A Direct Extension of SLSA~\cite{SLSA} to Graph Topology and Attributes}

An ideal graph of data points with $K$-classes should be nonnegative, symmetric and positive semi-definite with rank $K$ and have at most $\eta = \sum_k n_k^2$ nonzero entries, where $\{n_k\}$ are the sizes of the classes. Motivated by the structure of ideal graphs, the SLSA method \cite{SLSA} aims to refine a given matrix $A$ of $n$ samples or objects $\{x_i\}$ to a sparse and symmetric matrix $Z$ near $A$ such that the refined graph $Z$ has at most $\eta$ of nonzero entries with an estimated number $\eta$, and it approximates a positive semi-definite with rank equal to the class number $K$. The sparsity level $\eta$ of $Z$ is roughly estimated, depending on $K$ and $n$. 

SLSA solves the sparse problem
\begin{equation}\label{prob:slsa}
\begin{array}{c}
	\min_{Z,U} \big\{f(Z;A)+\theta\|Z-UU^{\top}\|_F^2\big\},
    \quad {\rm s.t.}\quad Z\geq 0,\ \|Z_{\text{off}}\|_0\leq \eta,\ U^{\top}U = I_{K}
\end{array}
\end{equation}
with a loss function $f(Z;A)$ of $Z$ to $A$ to obtain an refined graph $Z$, where $\|Z_{\text{off}}\|_0$ is the number of nonzero off-diagonals in $Z$. In this model, the orthonormal matrix $U$ has $K$ columns, and hence, the approximation between $Z$ and $UU^{\top}$ with a small error $\|Z-UU^{\top}\|_F^2$ implicitly imposes the constraints of positive semi-definition and rank $K$ on the refined matrix $Z$. Furthermore, due to the explicit sparse structure of $Z$, the approximation of $UU^{\top}$ to $Z$ also asks a clustered structure of the rows of $U$, as that in the spectral methods for graph cutting. The sparsity of $Z$ is directly imposed by the limit of its nonzero entries: $\|Z_{\text{off}}\|_0\leq \eta$, where $\|\cdot\|_0$ defines the off-diagonal cardinality of a matrix.

For computation, $A$ should be previously rescaled as 
$A:=\frac{\sqrt{K}}{\|A\|_F}A$to match $Z$, or equivalently $UU^\top$ in magnitude since $\|UU^\top\|_F = \sqrt{K}$. The loss function could be $f_1 = \|Z-A\|_1$ or $f_2=\|Z-A\|_F^2$, or the sparse forms
\[
    f_1^{(s)} = \sum_{(i,j)\in\Omega}|z_{ij}-a_{ij}|,\
    f_2^{(s)} = \sum_{(i,j)\in\Omega}(z_{ij}-a_{ij})^2,
\]
where $\Omega$ is the intersection of index sets of nonzero $Z$ and $A$. 
The sparse forms are more suitable to a relative dense graph $A$.
$\eta$ is an estimated integer of the off-diagonal cardinality of an ideal graph of the latent classes, for example, $\eta = K(\frac{n}{K})^2-n$, where $\frac{n}{K}$ is the average size of the $K$ classes. SLSA is not sensitive to the estimated sparsity level $\eta$.  

SLSA works well when the adjacency matrix $A$ has relatively evident class information. It can also help to increase the efficiency of many clustering algorithms for data analysis such as nonlinear dimensional reduction, unsupervised learning, or multi-view learning. However, it is expected that SLSA may lose its efficiency if the class information of the adjacency matrix is very weak as reported in \cite{SLSA}.

The attribute vectors $X = \{x_i\}$ equipped in the graph nodes of $A$ may provide auxiliary information of classes. Combining the auxiliary information may improve the adjacency $A$. For instance, construct a graph $G(X)$ from $X$ and combine it with $A$ by a convex combination of $A$ and $G(X)$,
\begin{align}\label{A}
    A_t = t \frac{A}{\|A\|_F}+(1-t)\frac{G(X)}{\|G(X)\|_F}
\end{align} 
with a suitable parameter $t\in [0, 1]$. Then,  apply SLSA on $A_t$, together with optimizing $t$ in the SLSA model. This idea leads to a slightly modified SLSA model
\begin{equation}\label{prob:slsa_t}
\begin{array}{c}
	\min_{Z,U,t} \big\{f(Z;A_t)+\alpha\|Z-UU^{\top}\|_F^2\big\},
     {s.t.} Z\!\geq\! 0,\ \|Z_{\text{off}}\|_0\!\leq\! \eta,\ U^{\top}U \!=\! I_{K},\ 0\!\leq\! t\!\leq\! 1
\end{array}
\end{equation}
for refining the weak graph $A$.

The matrix $G(X)$ could be a neighborhood graph of the attribute vectors $\{x_i\}$ without any structures of $X$, as in spectral projection methods.
Under the assumption that node attributes approximately lie in a union of low-dimensional subspaces, subspace learning techniques can be employed to extract the class information in the two steps: (a) 
obtain a coefficient matrix $C$ by solving the following optimization problem:
\[
    \min_{C} \|X - XC\|_F^2 + \zeta\|C\|_1,
\]
and (b) construct the attribute-induced graph 
    $G(X) = \frac{1}{2}\big(|C| + |C|^\top\big)$.
This approach is denoted as SLSA(t) in the numerical experiments reported later.

\subsection{Framework of Cross-Modality Learning (CML)}

SLSA(t) is a two-phase approach, directly combining the techniques for graph refinement and subspace learning. In this work, we aim to generalize the SLSA by combining the key idea of self-representation in subspace learning with the technique of SLSA in a more effective way. We assume
that 
\begin{enumerate}
    \item either the sparse adjacency matrix $A$ or the attribute matrix $X$ is highly noisy,
    \item the attribute points approximately lie on distinct subspaces that match the classes behind the adjacency matrix.
\end{enumerate} 

Existing joint graph learning methods usually adopt a fixed, unilateral mapping pipeline. When the anchored data domain is inherently noisy, these rigid models inevitably suffer from severe performance decay.

To break this learning barrier, we propose
a framework of Cross-Modality Learning (CML) for robust clustering based on a heavily noisy graph and the node-equipped attributes.
The core idea of CML is to treat joint subspace learning not as a static one-way pipeline, but as a flexible control system driven dynamically by data quality. By introducing cross-domain alignment within a concise three-term meta-model, our framework routes information bidirectionally: forward information propagation for graph topology recovery, and backward regularizing propagation for feature representation denoising.

To achieve this bidirectional mechanism, we organize the global objective function of CML into three explicit functional components:
\begin{itemize}
    \item \textbf{Topology Domain Regulating Term ($\mathcal{M}_{\text{Top}}$):} This component governs the formation of the target graph $Z$. It manages the proximity to the empirical graph template $A$ and applies spectral graph theory via an orthogonal indicator matrix $U$ to carve out a clean, block-diagonal clustering skeleton for $Z$ as SLSA.
    
    \item \textbf{Attribute Domain Regulating Term ($\mathcal{M}_{\text{Attr}}$):} This component constrains 
    a self-representation matrix $C$ of $X$. It captures the continuous manifold structure from the raw features $X$ via a secondary indicator matrix $V$ to retrieve feature-level clusters.

    \item \textbf{Cross-Domain Manifold Alignment Term:} This term acts as the bridge connecting the two distinct fields. By utilizing a continuous projection operator $\mathcal{T}(\cdot)$, it forces the algebraic self-representation matrix $C$ and the geometric target graph $Z$ to align in a shared space, ensuring seamless cross-domain synergy.
\end{itemize}

By assembling these three physical building blocks, we formalize the global objective function of the CML framework as follows:
\begin{align}
    \min_{C,Z,U,V} \quad & \mathcal{M}_{\text{Top}}(Z, A, UU^\top) + \mathcal{M}_{\text{Attr}}(X, C, VV^\top) + \gamma \|Z - \mathcal{T}(C)\|_F^2 \\
    \text{s.t.} \quad & Z \geq 0, \!\quad \|Z_{\rm off}\|_0 \leq \eta, \!\quad U^\top U = I_K,\! \quad V^\top V = I_K,\nonumber
\end{align}
where the constraints are the same as SLSA. 

In the following subsections, 
we demonstrate how this generalized framework naturally degenerates into two symmetric operational models based on data fidelity.

\subsection{Attribute-driven CML (A-CML)}

In this section, we consider the scenario where the observed graph $A$ is heavily corrupted, while the node attributes $X$ are relatively clean and reliable.

For the highly noisy graph topology, the topology domain focuses on recovering the graph template $A$ under a low-rank manifold constraint:
\begin{equation}
   \mathcal{M}_{\text{Top}}(Z, A, UU^\top)= \|Z-A\|_F^2+\beta\|Z-UU^\top\|_F^2,
\end{equation}
with $Z \geq 0$,$\|Z_{\rm off}\|_0 \leq \eta$ and $U^\top U=I_K$. This 
is the objective function of  the graph refinement approach SLSA. Since we assume $A$ is heavily corrupted, $Z$ cannot be accurately recovered by relying on the topology domain alone in this case. The relatively clean attribute matrix $X$ must be leveraged to guide and improve this recovery process.

For the attribute domain, when the features are noise-free, previous studies \cite{SSC,Liu2010,Liu2012} pointed that the self-representation matrix $C$ for $X = XC$ should have a rank equal to the sum of dimensions of these subspaces that the feature points lie to guarantee the subspace recovery, and the diagonal entries of $C$ should be nonzero but relatively small to guarantee the connection of $C$ within class \cite{minimal2019}. Theoretically, in the noise-free case, the representation matrix $C$ for the self-representation $X = XC$ is not unique and each feasible $C$ has the representation \cite{minimal2019}
\begin{align}\label{C}
    C(T) = VV^\top+V_\bot T  
\end{align}
for an arbitrary matrix $T$, 
where $V$ is an orthogonal basis matrix of the row space of $X$, giving $X = X(VV^\top)$, and $V_\bot$ is an orthogonal complement of $V$. 
The basic technique proposed in \cite{minimal2019} for subspace learning looks for a representation matrix $C(T)$ by
solving the problem
\begin{align}\label{subspace}
    \min_{T,\Omega} \|\Omega \odot C(T)\|_1+\frac{\lambda}{2}\|{\rm diag}(C(T))\|_2^2,    
\end{align}
where ${\rm diag}(C(T))$ is the vector of diagonals of $C(T)$, 
$\Omega$ is a indicator matrix of an estimated index set between different classes
and $\odot$ means the Hadamard product of two matrices. 

We integrate subspace learning into the SLSA model (\ref{prob:slsa}), 
taking two modifications on 
(\ref{subspace}). First, we relax the exact representation $C(S)$ in (\ref{C}) to an approximate $C$ with {an error $\|X-XC\|_F^2$ as small as possible. Second, we relate the indication matrix of active set to its implicit restriction by the penalty term $\|C-VV^\top\|_F^2$. That is, 
\begin{align}\label{Mattr}
   \mathcal{M}_{\text{Attr}}(X, C, VV^\top) 
   = &\, \alpha\, \|X-XC\|_F^2+\beta\|C-VV^\top\|_F^2 + \lambda\|\diag(C)\|_F^2,
\end{align}
subject to $V^\top V= I_K$. The third term penalized by $\lambda$ is used to suppress the trivial self-representation where each node simply reconstructs itself.

Finally, for the cross-domain alignment, 
because the empirical graph $A$ is heavily corrupted, a good subspace representation matrix $C$ must act as a "graph builder". 
Since $Z$ should be nonnegative and symmetric, we make use of the nonnegative symmetric form $S_+(C)=\frac{1}{2}\big(|C|+|C|^\top\big)$ of solution $C$ to push forward and actively reconstruct the graph space. Note that, in SLSA, the input matrix $A$ has been rescaled to have norm $\|A\|_F = \sqrt{K}$, equal to the norm of $\|UU^\top\|_F$. $Z$ is implicitly scaled as the same norm in magnitude as $UU^\top$. 

These two objective functions are linked by a penalty term like $\|Z-\mathcal{T}(C)\|_F^2$ with $\mathcal{T}(C) = sS_+(C)$, where $s$ balances the scales of $Z$ and $S_+(C)$, which leads to the Attribute-driven CML (A-CML) model:
\begin{align}\label{A-CML}
    \min_{C,Z,U,V}
    &  \{\|Z-A\|_F^2+\beta\|Z-UU^\top\|_F^2
        +\gamma\|Z-sS_+(C)\|_F^2 +\alpha \|X\!-\!XC\|_F^2 + \beta \|C-VV^\top\|_F^2 \notag\\
        & +\lambda\|\diag(C)\|_F^2\}, \quad
        { s. t.}\ Z\geq 0,\ \|Z_{\rm off}\|_0\leq \eta,\
        U^\top U = I_K,\ V^\top V = I_K.\!\!
\end{align}
Here, the model utilizes relatively high-quality features $X$ to generate a robust coefficient matrix $C$. Through the alignment term, these feature relationships are pushed forward into the graph domain to actively repair missing links and prune spurious edges in the corrupted template $A$.

\subsection{Topology-driven CML (T-CML)}

We now consider the branch of CML: Topology-driven CML (T-CML), which addresses a scenario opposed to that of A-CML. In this regime, the node attributes $X$ are assumed to be severely contaminated by environmental noise, whereas the empirical graph $A$ serves as a reliable prior topology.
This scenario is highly common in real-world applications, as collecting high-dimensional attributes is often more error-prone and complex than recording structural connections.

Under this uneven data quality, letting the heavily noisy features $X$ guide the graph structure as done in A-CML would mislead the model and propagate errors into the target graph. To avoid this, T-CML uses a backward denoising mechanism, where the relatively reliable structural information from $A$ is passed backward into the attribute space, acting as a geometric guide to clean up the feature representation noise. As a result, the three terms of our CML framework are designed as follows.

First, for the topology domain regulating term, since the empirical graph $A$ is relatively reliable, the target graph $Z$ is forced to be close to $A$:
\begin{equation}
    \mathcal{M}_{\text{Top}} = \|Z - A\|_F^2.
\end{equation}
Different from A-CML, we omit the low-rank constraint $\|Z - UU^\top\|_F^2$ here, because the high-quality graph $A$ already provides a relatively reliable cluster structure for $Z$.

Second, for the attribute domain regulating term, similarly to A-CML, the T-CML model also uses a subspace self-representation to capture feature structures, guided by an orthogonal matrix $V$:
\begin{equation}
    \mathcal{M}_{\text{Attr}} = \alpha \|X - XC\|_F^2 + \beta \|C - VV^\top\|_F^2 + \lambda \phi(C),
\end{equation}
where $\phi(C)$ is the self-loop normalization to avoid trivial self-reconstruction. Unlike A-CML, which uses dual-layer low-rank constraints on both $Z$ and $C$ to handle a noisy graph, T-CML places the low-rank clustering constraint only on the coefficient matrix $C$ to directly clean up the heavy noise in features $X$.

Third, for the cross-domain alignment term, different from A-CML, the goal shifts from fixing a bad graph to cleaning up noisy features. Since the empirical graph $A$ is reliable but the features $X$ are noisy, we drop the complex forward operator $\mathcal{T}(\cdot)$ and use a direct identity mapping. This allows the clean structure from the topology domain to flow backward and filter out noise in the feature representation matrix $C$. Driven by this mechanism, the alignment term forces direct proximity between the continuous matrix $C$ and the discrete matrix $Z$:
\begin{equation}
    \mathcal{M}_{\text{Align}} = \|C - Z\|_F^2.
\end{equation}
Through this formulation, the target graph $Z$, which matches the reliable graph $A$, pulls the noisy representation $C$ to follow the correct cluster boundaries. Furthermore, because the non-negativity of $Z$ passes directly to $C$ through this alignment, we can safely use the simple trace form $\phi(C)=2\text{tr}(C)$ without extra non-negative constraints on $C$.
By assembling these components, we arrive at the formulation of the T-CML model:
\begin{align}
    \min_{Z,C,V}  & \{\alpha\|X-XC\|_F^2 + \beta \|C-VV^\top\|_F^2 +2\lambda tr(C)  
     + \|Z-A\|_F^2 +\gamma \|C-Z\|_F^2\}  \nonumber\\
    \text { s.t. }  &Z\geq 0, {\|Z_{\text{off}}\|}_0 \leq \eta, V^{\top}V=I_K.
    \label{model:trcs_initial}
\end{align}

Although Eq. (\ref{model:trcs_initial}) faithfully fits our CML framework, jointly optimizing $\|Z-A\|_F^2$ and $\|C-Z\|_F^2$ introduces a subtle computational latency during alternating optimization iterations. Geometrically, in the matrix manifold defined by the Frobenius norm, minimizing the distance gaps in both $\|Z-A\|_F$ and $\|C-Z\|_F$ implicitly caps the upper bound of $\|C-A\|_F$ via the triangle inequality (i.e., $\|C-A\|_F \leq \|C-Z\|_F + \|Z-A\|_F$).  
To simplify the optimization, we rewrite the formulation into a simplified problem:
\begin{align}
    \min_{Z,C,V} \big\{ & \alpha\|X-XC\|_F^2 + \beta \|C-VV^\top\|_F^2  
    + 2\lambda tr(C) + \|C-A\|_F^2 +\gamma \|C-Z\|_F^2 \big\} \nonumber\\
    \text { s.t. }  &Z\geq 0, {\|Z_{\text{off}}\|}_0 \leq \eta, V^{\top}V=I_K.  
    \label{model:trcs}
\end{align}

\section{Algorithms}
\subsection{Solution of the A-CML model}
The minimization problem (\ref{A-CML}) can be solved by alternately optimizing one of the variables $C,Z,U,V$, each of the subproblems has a closed-form solution shown below.   

\textbf{Update $U$ and $V$ given $Z$ and $C$:}
The subproblem of optimizing $U$ is 
\begin{equation}
    \min_U \|Z-UU^\top\|_F^2 , \quad   {\rm s. t.}\ U^\top U = I_K,
    \label{eq:A-CML_U}
\end{equation}
while optimizing $V$ is equivalent to solving
\begin{equation}
    \min_V \|C-VV^\top\|_F^2 , \quad   {\rm s. t.}\ V^\top V = I_K,
    \label{eq:A-CML_V}
\end{equation}
similarly. Optimal solutions of these two problems are given by the $K$ eigenvectors corresponding to $K$ largest eigenvalues of $Z$ and the symmetric part ${\rm sym}(C)=(C+C^\top)/2$ of $C$, respectively. In the form of eigen-equation,
\begin{equation}\label{eq:Ueig}
    ZU = U\Lambda_Z,\quad 
    {\rm sym}(C)V = V\Lambda_C \notag
\end{equation}
with the diagonal matrices $\Lambda_Z$ and $\Lambda_C$ of the eigenvalues of $Z$ and ${\rm sym}(C)$, respectively.

\textbf{Update $C$ given $Z$:} The subproblem of optimizing $C$ is 
\begin{align*}
    \min_C &\{ \gamma\|Z-sS_+(C)\|_F^2+\alpha \|X-XC\|_F^2 + \beta \|C-VV^\top\|_F^2 +\lambda\|\diag(C)\|_F^2\}.    
\end{align*}
Since $C$ is approximately symmetric, 
\[
    \|Z-sS_+(C)\|_F \approx \|Z-s|C|\|_F 
    = \|Z-s C\odot{\rm sgn}(C)\|_F.
\]
We release the sign function of $C$ by the sign function of a previously estimated matrix of $C$, which yields
\begin{align}\label{eq:A-CML_C}
    \min_C &\{ \gamma\|Z-sC\odot{\rm sgn}(C)\|_F^2
        +\alpha \|X-XC\|_F^2  + \beta \|C-VV^\top\|_F^2 +\lambda\|\diag(C)\|_F^2\}. 
        \notag
\end{align}
Slightly modifying the objective function above by fixing the sign matrix ${\rm sgn}(C)$ to ${\rm sgn}(C_{\rm prev})$  for simplifying computation, its KKT condition for optimal solutions is
$$GC +\lambda \diag(C) = W,$$
where $G = \alpha X^\top X+ (\beta+\gamma s^2)I$ and $W =\alpha X^\top X + \beta VV^\top+\gamma s Z\odot{\rm sgn}(C_{\rm prev})$.
Let $M =G^{-1}, N =G^{-1}W$. This equation is equivalent to 
$$C + \lambda M\diag(C) = N.$$
Since the diagonals of $C$ can be previously determined by diagonals $m_{ii}$ and $n_{ii}$ of $M$ and $N$ as $c_{ii} = \frac{n_{ii}}{1+\lambda m_{ii}}$,
\begin{equation}
    C = N - M\diag\Big(\frac{n_{11}}{\lambda^{-1}+ m_{11}},\dots,\frac{n_{nn}}{\lambda^{-1}+m_{nn}}\Big).
    \label{eq:solutionC}
\end{equation}

\textbf{Update $Z$ given $U$ and $C$:}
When $C$ fixed,  we compute $S_+(C)$ and set $s = \frac{\sqrt K}{\|S_+(C)\|_F}$, the optimization of $Z$,
\begin{align}\label{eq:A-CML_Z}
    \min_{Z}
    &\ \big\{ \|Z-A\|_F^2+\beta\|Z-UU^\top\|_F^2
    +\gamma\|Z-sS_+(C)\|_F^2\} 
    {\quad s. t.} Z\geq 0,\ \|Z_{\rm off}\|_0\leq \eta
\end{align}
is an application of SLSA with a combination of $\|Z-A\|_F^2$ and $\|Z-sS_+(C)\|_F^2$. The optimal nonnegative and sparse $Z$ is component-wisely determined as follows. At first, we determine the optimal solution without the sparse and nonnegative restriction as
\[
    \hat{Z} = \frac{(A+\gamma sS_+(C)+\beta UU^\top)}{1+\beta+\gamma}.
\]
If the sparsity constraint is not taken into account,  \eqref{eq:A-CML_Z} has the optimal solution with entries
$z_{ij}=\hat z_{ij}^+=\max(\hat y_{ij},0).$
The objective value is given by $\phi_{ij}(\hat z_{ij}^+)= \max(-{\rm sgn}(\hat z_{ij})\hat z_{ij}^2,0)$. 
The sparsity constraint ${\|Z_{\text{off}}\|}_0 \leqslant \eta$ requires choosing some values of $\hat z_{ij}^+$ to vanish, which could cause a minimal increase of the corresponding value in the objective function. Specifically, reducing $\hat z_{ij}^+$ to $0$ will increase $\phi_{ij}(\hat z_{ij}^+)$ to $\phi_{ij}(0)$ with the gap
$$\phi_{ij}(0)-\phi_{ij}(\hat z_{ij}^+)=\max({\rm sgn}(\hat z_{ij})\hat z_{ij}^2,0)=(\hat z_{ij}^+)^2,\quad i\neq j.$$
Therefore, the optimal strategy is to set those $\hat z_{ij}^+$ to zero that correspond to the smallest gaps, while retaining those with the largest. 
If $\hat Z^+ = (\hat z^+)$ is $\eta$-separable, that is, its $\eta$-th largest value is strictly larger than $(\eta+1)$-th largest value, the optimal support set $\Omega_\eta$ for $Z$ can be chosen as the indices of the $\eta$ largest off-diagonal values in $\hat Z^+$. The optimal $Z$ is then obtained via a truncation operator $T_\eta$:
\begin{equation}
    Z=T_\eta(\hat Z^+) =
    \begin{cases}
    \hat z_{ij}^+ & \text{if }{(i,j)}\in \Omega_\eta;\\
    0 & \text{otherwise.} 
    \end{cases}
    \label{eq:T_eta}
    \end{equation}
To sum up, the closed-form solution of $Z$ of \eqref{eq:A-CML_Z} is
\begin{equation}
    Z=T_\eta\Big(\frac{\big(A+\beta UU^\top+\gamma s S_+(C)\big)^+}{1+\beta+\gamma}\Big).\label{updateZ}
\end{equation}

\begin{algorithm}[t]
\caption{Attribute-driven CML (A-CML)}
\label{alg:A-CML}
\textbf{Input}: symmetric adjacency matrix $A\in\mathbb{R}^{n\times n}$, feature matrix $X\in\mathbb{R}^{d\times n}$, group number $K$, parameter $\alpha,\beta,\gamma,\lambda,\eta$, tolerance $\tau$, max iterations $\max_{iter}$ \\
\textbf{Output}: $Z$, solution of Eq. \eqref{A-CML}
\begin{algorithmic}[1]
\STATE {Preprocessing:}
$A := \frac{\sqrt{K}}{\|A\|_F}A$ and $X := \frac{\sqrt{K}}{\|X\|_F}X$.

\STATE {Initialization:}
$Z = 0$, $C = A$, $s = 1$. 
\FOR{$k=1,\dots,\max_{iter}$}
    \STATE Set $Z_{\rm prev} = Z$;
    \STATE \textbf{Update $V$ and $C$:}
    \STATE Compute $V$ by the top-$K$ eigenvectors of $\frac{C+C^\top}{2}$;
    \STATE $W = \alpha X^\top X + \beta VV^\top + \gamma s (Z \odot \mathrm{sign}(C))$;
    \STATE $M = (\alpha X^\top X + (\beta + \gamma s^2) I)^{-1}$,\quad $N = MW$;
    \STATE Compute $c_i=\frac{N_{ii}}{1+\lambda M_{ii}}, C = N- \lambda M\mathrm{diag}\!\left(c_1,\cdots,c_n
    \right);$
    \STATE Compute $S_+(C) = (|C| + |C|^\top)/2$.
    \STATE \textbf{Update $Z$ and $U$:}
    \STATE Compute $U$ by the top-$K$ eigenvectors of $Z$;
    \STATE Update $Z=T_\eta\Big(\frac{(A+\beta UU^\top+\gamma s S_+(C))^+}{1+\beta+\gamma}\Big).$
    \STATE \textbf{Update scale $s$:}
    \quad $s = \sqrt{K}/\|S_+(C)\|_F$
    \STATE \textbf{Stopping check:} if $\|Z - Z_{\rm prev}\|_F \le \tau$, break
\ENDFOR
\RETURN $Z$
\end{algorithmic}
\end{algorithm}

\subsection{Solution of the T-CML model}
The optimization problem \eqref{model:trcs} could be solved by alternating iterations which optimize $V$, $C$ and $Z$ alternately in the objective function 
\begin{align}
F_t(Z,C,V) = &{\|C-A\|}_F^2 + \alpha {\|X-XC\|}_F^2 +  \beta{\|C-VV^{\top}\|}_F^2 + \gamma \|Z-C\|_F^2 + 2\lambda tr(C)  \notag
\end{align}
through the following three sub-problems:
\begin{subequations}
\begin{align}
    V_{k+1}&=\arg \min _V F_t\left(Z_{k}, C_{k}, V\right)  \quad  \text { s.t. } V^{\top} V=I_K   \label{eq:subV}\\
    C_{k+1}&= \arg \min _C F_t\left(Z_{k}, C, V_{k+1}\right)     \label{eq:subC}\\
    Z_{k+1}  &= \arg \min _Z F_t\left(Z, C_{k+1}, V_{k+1}\right) \ \ 
    \text { s.t. }  Z \geq 0, \left\|Z_{\text{off}}\right\|_0 \leq \eta. 
    \label{eq:subZ}  
\end{align}
\end{subequations}

The first subproblem \eqref{eq:subV} is to solve 
\begin{equation}
    \min_{V^{\top} V=I_K} \|C_k-VV^{\top}\|_F^2,
    \label{eq:subopt1}
\end{equation} 
where $C_k$ is initially set as $C_0=A$. Similar to \eqref{eq:A-CML_U}, its solution is given by the top $K$ eigenvectors of $C_{k}$ corresponding to the $K$ largest eigenvalues. 

The second subproblem \eqref{eq:subC} is to solve the unconstrained problem
\begin{align}
\min_{C}&\ \{{\|C-A\|}_F^2 + \alpha {\|X-XC\|}_F^2 +  \beta{\|C-Q_{k+1}\|}_F^2  +\gamma \|Z_k-C\|_F^2 + 2\lambda tr(C)\},\notag
\end{align}
where $Q_{k+1}=V_{k+1}V_{k+1}^{\top}$. The optimal solution can be obtained by gradient descent algorithm based on the gradient
\begin{align}
&\nabla_C F_t(Z_k,C,V_{k+1})=2[(1+\beta+\gamma)I+\alpha X^{\top}X]C -2(A+\alpha X^{\top}X +\beta Q_{k+1}+\gamma Z_k-\lambda I), \notag
\end{align}
or by the closed form solution $\nabla_C F_t(Z_k,C,V_{k+1})=0$ with
\begin{align}
    C_{k+1} = \left[(1+\beta+\gamma)I+\alpha X^{\top} X\right]^{-1} ( A+\alpha  X^{\top} X + \beta Q_{k+1}  + \gamma Z_k - \lambda I)
     \label{eq:C}
\end{align} 

The third subproblem \eqref{eq:subZ} is to solve the optimization problem:
\begin{equation}
    \min_Z \|Z-C_{k+1}\|_F^2 \quad \text{s.t. } Z\geq 0, {\|Z_{\text{off}}\|}_0 \leq \eta. 
    \label{eq:subopt3}
\end{equation}
Similar to \eqref{eq:A-CML_Z}, the optimal solution is given by
\begin{equation}
    Z_{k+1}=T_\eta(C_{k+1}^+) =
    \begin{cases}
    {c}_{ij}^+ & \text{if }{(i,j)}\in \Omega_\eta;\\
    0 & \text{otherwise .} 
    \end{cases}
    \label{eq:Z}
    \end{equation}  
provided $C_{k+1}^+$ is $\eta$-separable.

\begin{algorithm}[t]
\caption{Topology-driven CML (T-CML)}\label{alg:T-CML}
\textbf{Input}: 
symmetric adjacency matrix $A\in\mathbb{R}^{n\times n}$, feature matrix $X\in\mathbb{R}^{d\times n}$, group number $K$, parameter $\alpha,\beta,\gamma,\eta$, tolerance $\tau$, max iterations $\max_{iter}$ \\
\textbf{Output}: $Z$, solution of Eq. \eqref{model:trcs}
\begin{algorithmic}[1]
\STATE   Filter $X$ and rescale $A := \frac{\sqrt{K}}{\|A\|_F}A$ and $X := \frac{\sqrt{K}}{\|X\|_F}X$.
\STATE  {Initialization:} $Z = 0, C=A$. 
\FOR{$k=1,\dots,\max_{iter}$}
    \STATE   Compute $K$ largest eigenvectors $\{ v_i\}$  of $C_k$  and set $V_{k+1} =[v_1, \dots, v_K]$.
    \STATE  Compute the dense graph $C_{k+1}$ by Eq. \eqref{eq:C}. 
    \STATE   Compute ${Z}_{k+1}$ as Eq. \eqref{eq:Z}.
    \STATE   Set $k=k+1$. 
    \IF{$\|Z_{k+1}-Z_k\|_F/\|Z_k\|_F < \tau$} 
        \STATE break
        \ENDIF
\ENDFOR
\STATE  \textbf{Return $Z=Z_k$}     
\end{algorithmic}
\end{algorithm}

\subsection{Fast-CML}

The computational complexity of CML is mainly due to the computation of high-dimensional matrix inverse in Equation \eqref{eq:C} and the truncation of the large matrix $C$ in Equation \eqref{eq:Z}. Here, for very large graphs, we propose a feasible large-scale method Fast-CML, which makes use of landmark nodes as inspired by the other large-scale approach in \cite{large}. 
The $m$ landmark nodes are first obtained by cluster centers of K-means on all samples, and $H\in \mathbb{R}^{d\times m}$ denotes the attribute features for these $m$ landmarks. 
For large-scale problem, we first denoise the subgraph with adjacency matrix $\hat{A}\in \mathbb{R}^{m\times n}$ which connects the whole $n$ nodes and the $m$ landmark nodes, with the following Fast-CML model,
\begin{align}
    &\min_{\hat{Z},U_1,U_2} {\|\hat{Z}-\hat{A}\|}_F^2 
    + \alpha {\|X-H\hat{Z}\|}_F^2 +  \beta{\|\hat{Z}-U_1U_2\|}_F^2, \quad \text {s.t. }  \hat{Z}\geq 0, \label{fast}
\end{align}
where $U_1\in \mathbb{R}^{m\times K}$, $U_2\in \mathbb{R}^{K\times n}$.
The optimal $\hat{Z} \in \mathbb{R}^{m\times n}$ clarifies the graph between the landmarks nodes and all nodes. Based on this subgraph, the refined whole graph $Z$ is reconstructed by $Z=\hat{Z}^{\top}\hat{Z}$. For node clustering, we use a Spectral Clustering-like strategy, which first computes the eigenvectors of $Z$ corresponding to its largest $K$ eigenvectors, and then applies $k$-means on these eigenvectors.
Note that, one doesn't need to compute $Z$ explicitly, since finding the eigenvectors of $Z$ is equivalent to computing the singular vectors of $\hat{Z}$.

The optimization \eqref{fast} could be similarly solved by alternating iterations, and in each iteration, we compute $U_1,U_2,\hat{Z},\hat{H}$ sequentially by:
\begin{equation}
\left\{
\begin{aligned}
U_1 &= \hat{Z} U_2^{\top} (U_2U_2^{\top})^{-1}\\
U_2 &= (U_1^{\top}U_1)^{-1} U_1^{\top}\hat{Z}\\
\hat{Z} &= \{((\beta+1)I_m+\alpha H^{\top}H)^{-1}
          (\hat{A}+\alpha H^{\top}X + \beta U_1U_2)\}_{+}\\
H &= X\hat{Z}^{\top}(\hat{Z}\hat{Z}^{\top})^{-1}.
\end{aligned}
\right.
\end{equation}
The computational complexity of the Fast-CML is 
\[ O(mnK)+O(mnK)+O(mnd)+O(mnd)=O(mnd).\]

\section{Computational Complexity}

In this section, we provide a computational complexity analysis
for both A-CML and T-CML. We
explicitly account for the number of dominant matrix operations, such as
low-rank multiplications ($\mathcal{O}(n^2K)$), feature-related products
($\mathcal{O}(n^2d)$), and sorting operations ($\mathcal{O}(n^2\log n)$), allowing a more accurate comparison of their practical efficiency. 

\subsection{Complexity of A-CML}

\textbf{Updating $U$, $V$.}
Computing the top-$K$ eigenvectors of $Z$ and the symmetric part of $C$ to obtain $U$ and $V$ requires two eigen-decompositions, with a complexity of $2 \cdot \mathcal{O}(n^2 K)$.

\textbf{Updating $C$.}
The update of $C$ involves several matrix operations. First, the matrix
$X^\top X$ is computed with a complexity of $\mathcal{O}(n^2 d)$, which can
be precomputed once before the iterations. 
Next, the low-rank matrix
$VV^\top$ is constructed with a cost of $\mathcal{O}(n^2 K)$, and the
Hadamard product $Z \odot \mathrm{sign}(C_{\rm prev})$ is evaluated with
complexity $\mathcal{O}(n^2)$. Based on these quantities, the matrices
$G = \alpha X^\top X + (\beta + \gamma s^2) I$ and
$W = \alpha X^\top X + \beta VV^\top + \gamma s Z \odot \mathrm{sign}(C_{\rm prev})$
are formed.

When the number of nodes $n$ is large, computing $M = G^{-1}$ resulting in a computational complexity of $O(n^3)$.
To handle large-scale data, from the well-known Woodbury matrix identity, we could obtain a cheaper way to compute the inverse:
\begin{equation}
\scalebox{1}{$
\begin{aligned}
    &\left[cI+\alpha X^{\top} X\right]^{-1} 
    =\frac{1}{c} I-\frac{1}{c^2} X^{\top}\left(\frac{1}{\alpha} I_d+\frac{1}{c} X X^{\top}\right)^{-1} X  
\end{aligned} $}
\end{equation}
This step mainly involves matrix multiplications resulting in a complexity of
$2 \cdot \mathcal{O}(n^2 d)$, together with the inversion of a
$d \times d$ matrix, which requires $\mathcal{O}(d^3)$ operations.
Subsequently, the matrix $N = MW$ is computed, incurring an additional
$\mathcal{O}(n^2 d)$ cost.

Finally, the solution of $C$ is obtained via a diagonal correction as \begin{equation}
    C = N - M\diag\Big(\frac{n_{11}}{\lambda^{-1}+ m_{11}},\dots,\frac{n_{nn}}{\lambda^{-1}+m_{nn}}\Big).
    \label{eq:solutionC}
\end{equation}
This step requires multiplying $M$ with a diagonal
matrix, which costs $\mathcal{O}(n^2)$. 

\textbf{Computing $S_+(C)$ and updating the scale $s$.}
The symmetric matrix $S_+(C)$ is computed as
$S_+(C) = (|C| + |C|^\top)/2$, which involves elementwise absolute value
and matrix addition operations with a total complexity of $\mathcal{O}(n^2)$.
The scale parameter is updated as
$s = \sqrt{K}/\|S_+(C)\|_F$, where the Frobenius norm
requires $\mathcal{O}(n^2)$ operations. 

\textbf{Updating $Z$.}
The update of $Z$ involves combining multiple dense matrices followed
by a truncation operation as \begin{equation}
    Z=T_\eta\Big(\frac{\big(A+\beta UU^\top+\gamma s S_+(C)\big)^+}{1+\beta+\gamma}\Big).
\end{equation}
Specifically, $Z$ is computed by first forming the low-rank matrix $UU^\top$, which requires
$\mathcal{O}(n^2 K)$ operations, and then aggregating it with $A$ and
$S_+(C)$ through matrix additions with cost $\mathcal{O}(n^2)$. Finally,
the truncation operator $T_\eta$ requires sorting the entries of the matrix and incurs a computational
cost of $\mathcal{O}(n^2 \log n)$. 

In summary, the per-iteration complexity of A-CML 
is
\[
    \mathcal{O}\big(n^2(K+d+1+\log n)\big).
\]

\subsection{Complexity of T-CML}

\textbf{Updating $U$.}
Computing the top-$K$ eigenvectors of $C_k$ requires $\mathcal{O}(n^2 K)$.

\textbf{Updating $C$.}
Similar to A-CML, using the Woodbury matrix identity, the matrix inversion in 
\begin{align}
    C_{k+1} = &\left[(1+\beta+\gamma)I+\alpha X^{\top} X\right]^{-1} ( A+\alpha  X^{\top} X + \beta Q_{k+1}  + \gamma Z_k - \lambda  I),
\end{align} 
 can be computed efficiently. The dominant cost comes from
matrix multiplications resulting in
a complexity of $2 \cdot \mathcal{O}(n^2d)$, together with the inversion of a
$d\times d$ matrix, which requires $\mathcal{O}(d^3)$ operations

\textbf{Updating $Z$.}
The update of $Z$ requires selecting the top-$\eta$ entries from
$C_{k+1}^+$, which involves sorting $\mathcal{O}(n^2 \log n)$.

In summary, the per-iteration complexity of T-CML is $\mathcal{O}\big(n^2(K+d+\log n)\big)$.

\section{Convergence Analysis}

We analyze the convergence properties of the proposed algorithm for A-CML. 
Specifically, we establish the convergence of the objective sequence $\{F(Z_k,C_k,U_k,V_k)\}$, 
the sequences $\{Z_k\}$ and $\{C_k\}$, as well as the orthogonal projection sequences 
$\{U_kU_k^\top\}$ and $\{V_kV_k^\top\}$. 
Moreover, we show that any accumulation point of $\{Z_k, C_k, U_k, V_k\}$ satisfies the optimality conditions under certain conditions. We denote the objective function of A-CML as
\begin{align}
&F_a(Z,C, U,V)= \|Z - A\|_F^2  +  \beta\|Z - UU^\top\|_F^2  +  \alpha \|X - XC\|_F^2  \nonumber\\
&+ \gamma\|Z - sS_+(C)\|_F^2  +  \beta \|C - VV^\top\|_F^2  + \lambda\|\diag(C)\|_F^2.
\end{align}
We first obtain the KKT conditions for the optimization problem \eqref{A-CML} in the following Lemma. 

\begin{lemma}
The KKT conditions of the optimization problem \eqref{A-CML}
are given as follows.
There exists a nonnegative matrix $\Sigma \geq 0$ such that
\begin{equation}
\Sigma = 2(Z-A) + 2\beta(Z-UU^\top) 
+ 2\gamma(Z - sS_+(C)), \label{eq:KKT_Z}
\end{equation}
with complementary condition $\Sigma \odot Z = 0$, and
\begin{align}
&0 \in 2\alpha X^\top(XC - X) 
+ 2\beta(C - VV^\top) 
+ 2\lambda(\diag(C))-2\gamma s \, \partial |C| \odot (Z - sS_+(C)), \label{eq:KKT_C}\\
&Z U = U \diag(\lambda_1(Z),\dots,\lambda_K(Z)), \label{eq:KKT_U}\\
&{\rm sym}(C) V = V \diag(\lambda_1(C),\dots,\lambda_K(C)),\label{eq:KKT_V}\\
& Z = T_\eta(\frac{(A+\beta UU^\top+\gamma s S_+(C))^+}{1+\beta+\gamma}), \label{eq:KKT_L0}
\end{align}
\end{lemma}

\begin{pf}
The feasible set induced by the sparsity constraint $\|Z_{\rm off}\|_0 \leq \eta$ 
can be represented as a finite union of subsets, each corresponding to a fixed support set of nonzero entries. Consequently, the analysis can be restricted to each subset individually, where the support is fixed. Under a given support, the problem reduces to a constrained optimization problem involving only nonnegativity and orthogonality constraints, for which the KKT conditions can be derived.

The Lagrangian function of the optimization problem is
\begin{align}
    L(Z,C,U,V,H_U,H_V,\Sigma) = F(Z,C,U,V) \!-\! \langle \Sigma,Z\rangle\! + \!\langle H,U^\top U \!-\!I_K\rangle\!+\!\langle J,V^\top V \!-\! I_K\rangle,  \label{eq:Lagrangianfunction}
\end{align}
where $H,J$ and $\Sigma \geq 0$ are Lagrange multipliers for the orthogonality and nonnegativity constraints, respectively.

\noindent
\textbf{(1) Optimality condition with respect to $Z$.}

Fixing $C,U,V$, the derivative of $L$ with respect to $Z$ gives
$$ \frac{\partial L}{\partial Z}
= 2(Z\!-\!A) \!+\! 2\beta(Z\!-\!UU^\top) \!+\! 2\gamma(Z \!-\! sS_+(C))\!-\!\Sigma=0.$$
Introducing the nonnegative multiplier $\Sigma \geq 0$ for the constraint $Z \geq 0$, 
the KKT condition is $\Sigma = 2(Z-A) + 2\beta(Z-UU^\top) 
+ 2\gamma(Z - sS_+(C))$, together with the complementary slackness condition $\Sigma \odot Z = 0$, 
which yields \eqref{eq:KKT_Z}.

\vspace{0.5em}
\noindent
\textbf{(2) Optimality condition with respect to $C$.}

The terms in the Lagrangian \eqref{eq:Lagrangianfunction} that depend on $C$ are
\begin{equation}
\alpha \|X\!-\!XC\|_F^2 \!+\! \beta \|C\!-\!VV^\top\|_F^2 \!+\!\lambda\|\diag(C)\|_F^2 \!+\! \gamma R(C), \notag
\end{equation}
where $R(C) = \|Z-sS_+(C)\|_F^2 =\sum_{i,j} \left(Z_{ij} - \frac{s}{2}\big(|C_{ij}| + |C_{ji}|\big)\right)^2.$
The subgradient of \(R(C)\) with respect to an entry \(C_{pq}\) can be derived by observing that \(C_{pq}\) contributes to both \(S_+(C)_{pq}\) and \(S_+(C)_{qp}\). More precisely, by the definition of \(S_+(C)\), we have
\[
\frac{\partial S_+(C)_{ij}}{\partial C_{pq}} = \frac{1}{2} \big( \delta_{ip}\delta_{jq} + \delta_{iq}\delta_{jp} \big) \cdot \partial |C_{pq}|,
\]
where \(\delta_{ij}\) is the Kronecker delta, equal to 1 if \(i=j\) and 0 otherwise. Here \(\partial |C_{pq}|\) is the sub-differential of the absolute value given by
\[
\partial |C_{pq}| = 
\begin{cases}
1, & C_{pq} > 0,\\
[-1,1], & C_{pq} = 0,\\
-1, & C_{pq} < 0.
\end{cases}
\]
Substituting this into element-wise derivative of \(R(C)\) gives
\begin{align}
&\frac{\partial R(C)}{\partial C_{pq}} \!=\! -\!2 s \sum_{i,j} (Z_{ij} \!-\! s S_+(C)_{ij}) \frac{\partial S_+(C)_{ij}}{\partial C_{pq}}\!=\! -s \Big[(Z_{pq} \!-\! s S_+(C)_{pq}) + (Z_{qp} \!-\! s S_+(C)_{qp})\Big] \!\cdot\! \partial |C_{pq}|. \notag
\end{align}
Finally, by the symmetry of \(Z\) and \(S_+(C)\), the sub-gradient can be compactly expressed in matrix form as
\[
\partial_C R(C) = -2 s (Z - s S_+(C)) \odot \partial |C|,
\]
where \(\partial |C|\) denotes the element-wise sub-differential of \(|C|\).
Combining this with the gradients of the smooth terms, the optimality condition with respect to \(C\) is
\begin{align}
0 \in\;& 2\alpha X^\top(XC - X) 
+ 2\beta(C - VV^\top) 
+ 2\lambda (\diag(C))-2\gamma s \, (Z - sS_+(C)) \odot \partial |C|, \notag
\end{align}
which coincides with \eqref{eq:KKT_C}.

\vspace{0.5em}
\noindent
\textbf{(3) Optimality condition with respect to $U$ and $V$.}

For $U$, the first-order optimality condition is
$$\nabla_U L=-4\beta(Z - UU^\top)U + 2U H=0,$$ 
where $H \in \mathbb{R}^{K \times K}$ is a symmetric matrix.
Rearranging the above equation, we get
\begin{align}
ZU = U\left(I_K + \tfrac{1}{2\beta}H\right) = U\tilde{H}, \notag
\end{align}
where $\tilde{H} = I_K + \tfrac{1}{2\beta}H$ is symmetric.
Hence, $\tilde{H}$ admits an eigen-decomposition $\tilde{H} = G \Lambda G^\top$, 
with orthogonal $G$ and diagonal $\Lambda = \diag(\lambda_1,\dots,\lambda_K)$. 
Letting $\hat{U} = UG$, we obtain $Z\hat{U} = \hat{U}\Lambda$, 
which shows that the columns of $\hat{U}$ are eigenvectors of $Z$. 
Therefore, $U$ consists of the $K$ largest eigenvectors of $Z$, yielding \eqref{eq:KKT_U}.

Similarly, for $V$, the optimality condition is
\[
\nabla_V L = -4\beta({\rm sym}(C)-VV^\top)V + 2V J = 0,
\]
with symmetric $\text{sym}(C)=(C +C^\top)/2$ and $J$.
Similarly, $V$ consists of the top-$K$ eigenvectors of $\it{sym}(C)$, yielding \eqref{eq:KKT_V}.

Finally, the truncation operator $T_\eta$ implicitly encodes the combinatorial sparsity constraint by selecting the optimal support, allowing the KKT analysis to be conducted on a fixed support set without explicitly handling the $\mathcal{L}_0$ sparse constraint. 
For each support, the optimal solution coincides with
the projection of an unconstrained solution onto the set of matrices
with at most $\eta$ nonzero off-diagonal entries. This projection can be
compactly expressed via the truncation operator $T_\eta$. Therefore, the
optimal $Z$ satisfies
\begin{equation}
Z = T_\eta(\hat{Z})=T_\eta(\frac{(A + \beta UU^\top + \gamma s S_+(C))^+}{1+\beta+\gamma}),
\notag
\end{equation}
where $\hat{Z}$ denotes a stationary point satisfying the KKT conditions of the problem without the sparsity constraint.

This completes the proof.
\end{pf}
\vspace{0.5cm}

\begin{theorem}\label{thm:conv}
Let $\{Z_k, C_k, U_k, V_k\}$ be generated by the algorithm A-CML, and let $P_k = U_kU_k^{\top}$, $Q_k = V_kV_k^{\top}$. Then the following statements are true.
\begin{itemize}
\item[(a)] The sequence $\{F_a(Z_k, C_k, U_k, V_k)\}$ converges.
\item[(b)] If $\lambda_K(Z_*)>\lambda_{K+1}(Z_*)$ for any accumulation point $Z_*$ of $\left\{Z_k\right\}$, the matrix sequence $\{P_k-P_{k+1}\}$ converges to zero.
Similarly, if $\lambda_K(C_*)>\lambda_{K+1}(C_*)$ for any accumulation point $C_*$ of $\left\{C_k\right\}$, $\{Q_k-Q_{k+1}\}$ converges to zero.
\item[(c)] If $\{P_k\}$ has an isolated accumulation point $P_*=U_*U_*^{\top}$ and $\lambda_K(Z_*)>\lambda_{K+1}(Z_*)$ for any accumulation point $Z_*$ of $\left\{Z_k\right\}$, $P_k$ converges to $P_*$. 
If $\{Q_k\}$ has an isolated accumulation point $Q_*=V_*V_*^{\top}$ and $\lambda_K(C_*)>\lambda_{K+1}(C_*)$ for any accumulation point $C_*$ of $\left\{C_k\right\}$, $Q_k$ converges to $Q_*$ and $\{C_k\}$ converges. 
Thus, $\{\hat{Z}_k\}$ converges.
Furthermore, if $\hat{Z}_*=\lim_{k\to\infty}\hat{Z}_k $ is $\eta$-separable, then $\{Z_k\}$ converges to $T_{\eta}(\hat{Z}_*^+)$.
\item[(d)] Any accumulation point $\left(Z_*,C_*,  U_*,V_*\right)$ of the sequence $\left\{\left(Z_k, C_k, U_{k+1},V_{k+1}\right)\right\}$ satisfies the KKT conditions of \eqref{A-CML} if $\lambda_K\left(Z_*\right)>\lambda_{K+1}\left(Z_*\right)$ and $\lambda_K\left(C_*\right)>\lambda_{K+1}\left(C_*\right)$.
\end{itemize}
\end{theorem}

\begin{pf} The whole proof consists of four parts corresponding to the four results.

\textbf{Part (a).}
The alternating optimization of subproblems \eqref{eq:A-CML_U}--\eqref{eq:A-CML_Z} ensures the monotonic decrease of the objective function:
\begin{align}
    &F(Z_{k}, C_{k}, U_{k}, V_k) \geq F(Z_{k}, C_{k}, U_{k+1}, V_k)\geq F(Z_{k}, C_{k},U_{k+1},V_{k+1}) \notag \\
    & \geq F(Z_{k}, C_{k+1},U_{k+1},V_{k+1})  \geq F(Z_{k+1}, C_{k+1},U_{k+1},V_{k+1}).
    \label{eq:decrese}
\end{align}
Since $F$ is non-negative and monotonically decreasing, its convergence is guaranteed.

\textbf{Part (b).}
At each iteration, $P_k$ and $Q_k$ are the top-$K$ eigenvectors of $Z_k$ and $C_k$, respectively. Following Lemma 4.1 in SLSA \cite{SLSA}, if any accumulation point $Z_*$ (or $C_*$) satisfies the spectral gap condition $\lambda_K(\cdot) > \lambda_{K+1}(\cdot)$, it follows that $P_k - P_{k+1} \to 0$ (or $Q_k - Q_{k+1} \to 0$).

\textbf{Part (c).}
Since the condition in (b) holds, we have $\{P_k-P_{k+1}\}\to 0$ and $\{Q_k-Q_{k+1}\}\to 0$.  Because $\{P_k\}$ and $\{Q_k\}$ have their isolated accumulation point $P_*$ and $Q_*$, 
by Lemma 4.10 of \cite{More1983}, if $P_k$ does not converge, there exists a subsequence $\{P_{k_j}\}$ of $\{P_k\}$ and an $\epsilon>0$ such that $\|P_{k_j+1}-P_{k_{j}}\|_F\geq \epsilon$,
which results in a contradiction with $P_k-P_{k+1}\to 0$. Therefore, $\{P_k\}$ converges to its accumulation point $P_*$. Similarly, $\{Q_k\}$ also converges to its accumulation point $Q_*$.
On the other hand, since $F(Z,C, U,V)$ is continuous, the convergence of $\{Q_k\}$ implies that the sequence $\{C_k\}$ also converges.
The convergence of $\{P_k\}$ and $\{C_k\}$ implies that the sequence $\{Z_k\}$ also converges. 
Since $\hat{Z}_*$ is $\eta$-separable, $Z_k$ is also $\eta$-separable for sufficiently large $k$, and the index set of $\eta$-largest elements in $\hat{Z}_k$ does not change for sufficiently large $k$. Thus, $ \{Z_k=T_{\eta}(\hat{Z}_k^+)\}$ converges to $T_{\eta}(\hat{Z}_*^+)$.

\textbf{Part (d).}
Lemma 1 shows the KKT conditions of \eqref{A-CML}. Let $(Z_*, C_*, U_*, V_*)$ be an accumulation point of the sequence $\{(Z_k, C_k, U_{k+1}, V_{k+1})\}$ with a convergent subsequence $\{(Z_{k_j}, C_{k_j}, U_{k_j+1}, V_{k_j+1})\}$. We denote $P_* = U_* U_*^\top$ and $Q_* = V_* V_*^\top$.

By the algorithm, $U_{k+1}$ consists of the $K$ leading eigenvectors of $Z_k$, and $V_{k+1}$ consists of the $K$ leading eigenvectors of $\frac{C_k + C_k^\top}{2}$.
Taking limits along the subsequence and using the spectral gap assumptions $\lambda_K(Z_*) > \lambda_{K+1}(Z_*)$ and $\lambda_K(C_*) > \lambda_{K+1}(C_*)$, yielding
$$Z_* U_* = U_* \operatorname{diag}(\lambda_1(Z_*),\dots,\lambda_K(Z_*)),$$
and $\frac{C_* + C_*^\top}{2} V_* = V_* \operatorname{diag}(\lambda_1(C_*),\dots,\lambda_K(C_*))$, which correspond to the KKT conditions \eqref{eq:KKT_U} and \eqref{eq:KKT_V}.

Next, consider the update for $C$. For fixed $Z_k, V_{k+1}$, the matrix $C_{k+1}$ minimizes
$\alpha\|X - XC\|_F^2 + \beta\|C - V_{k+1}V_{k+1}^\top\|_F^2 + \gamma\|Z_k - s S_+(C)\|_F^2 + \lambda\|\operatorname{diag}(C)\|_F^2$.
The corresponding first-order optimality condition ensures
\begin{align*}
0 \!\in\!   2\alpha X^\top (X C_{k+1} \!-\! X) \!+\! 2\beta (C_{k+1} \!-\! V_{k+1}V_{k+1}^\top) \!+\! 2\lambda\operatorname{diag}(C_{k+1}) 
\!-\! 2\gamma s \, \delta|C_{k+1}| \odot (Z_k \!-\! s S_+(C_{k+1})).
\end{align*}
By taking the limit along the convergent subsequence, and using the continuity of all smooth terms in $C_{k+1}$, $Z_k$ and $V_{k+1}$, together with the fact that the sub-differential of $|\cdot|$ is locally bounded (so we may extract a convergent sub-subsequence of the sub-gradients), we can obtain
\begin{align*}
0 &\in 2\alpha X^\top (X C_* - X) + 2\beta (C_* - Q_*) + 2\lambda\operatorname{diag}(C_*) - 2\gamma s \, \delta|C_*| \odot (Z_* - s S_+(C_*)),
\end{align*}
which is precisely the KKT condition \eqref{eq:KKT_C}.

Finally, for the $Z$-subproblem under nonnegativity alone, the exact solution is
\[
\hat{Z} = \frac{(A + \beta UU^\top + \gamma s S_+(C))^+}{1+\beta+\gamma},
\]
The nonnegativity constraint is enforced by the positive-part operation.
The KKT conditions for $\hat{Z}_k$ guarantee the existence of a multiplier $\Sigma_k \ge 0$ with $\Sigma_k \odot \hat{Z}_k = 0$ and 
\[
\Sigma_k = 2(\hat{Z}_k - A) + 2\beta(\hat{Z}_k - P_k) + 2\gamma(\hat{Z}_k - s S_+(C_k)).
\]
By the convergence of $\{P_k\}$ and $\{C_k\}$, $\{\hat{Z}_k\}$ converges to some $\hat{Z}_*$.
The sequence $\{\Sigma_k\}$ is bounded (the sub-gradients are bounded), 
so we may extract a convergent subsequence with limit $\Sigma_* \ge 0$ satisfying $\Sigma_* \odot \hat{Z}_* = 0$ and
\[
\Sigma_* = 2(\hat{Z}_* - A) + 2\beta(\hat{Z}_* - P_*) + 2\gamma(\hat{Z}_* - s S_+(C_*)).
\]
The actual update is $Z_k = T_\eta(\hat{Z}_k)$.
Because $\hat{Z}_*$ is $\eta$-separable, the truncation operator $T_\eta$ is continuous in a neighborhood of $\hat{Z}_*$.
Taking limits gives $Z_* = T_\eta(\hat{Z}_*)$, i.e.,
\[
Z_* = T_\eta\!\left( \frac{(A + \beta P_* + \gamma s S_+(C_*))^+}{1+\beta+\gamma} \right),
\]
which is the KKT condition \eqref{eq:KKT_L0}. Moreover, the multiplier $\Sigma_*$ can be restricted to the support of $Z_*$ so as to satisfy the complementarity condition exactly.

Overall, $(Z_*,C_*,U_*,V_*)$ satisfies the KKT conditions in \textit{Lemma 1}, and the proof is completed.
\end{pf}
\vspace{0.5cm}

The convergence of T-CML is similar. Its KKT conditions are as follows. We give the following lemma and theorem to show the convergence of the sequence $\{F_t(Z_k,C_k, V_k)\}$ in T-CML, convergence of the sequence of orthogonal projections $\{Q_k\}$ and $Z_k$, and the optimality of convergence points of $\{Z_k, C_k, V_k\}$.

\begin{lemma}
The KKT conditions of T-CML are that there is a nonnegative matrix $\Sigma$ such that
\begin{equation}
\begin{cases}       
\Sigma = 2[C-A+\alpha X^{\top}(X-XC) + \beta(C-{V}{V}^{\top})]; \\ 
CV = V\Lambda;\\
\Sigma \odot Z=0;\\
Z=T_\eta(C^+).  
\end{cases} \notag
\end{equation}
\end{lemma}

\begin{pf}
The feasible solutions of T-CML Eq.\eqref{model:trcs} can be classified into a finite number of sets for the sparse restriction $\|Z_{\text{off}}\|_0 \leq\eta$. Each set has a fixed support set of the nonzero entries of the solutions. 
Therefore, the KKT conditions should be the KKT conditions subjected to each support set. This implies that the sparse restriction $\|Z_{\text{off}}\|_0 \leq\eta$ can be replaced by restricting the entries of $Z$ in a fixed support set of size $\eta$.
Therefore, we first consider the convex constraints of T-CML \eqref{model:trcs}. Constraint $Z \geq 0$ can be divided into $n^2$ constraints $z_{ij} \geq 0(i,j=1,\dots,n^2)$ , thus we need $n^2$ multipliers subject to $\sigma_{ij}\geq 0, \sigma_{ij}z_{ij}=0$ which can be equivalently replaced by $\Sigma \geq 0$ and $\Sigma \odot Z=0$.
The lagrange function without sparse constraint is:
\begin{align}
    L=& \|C-A\|_F^2 + \alpha \|X-XC\|_F^2 + \beta \|C-VV^{\top}\|_F^2 + \notag\\
    & \gamma \|C-Z\|_F^2 + <H,V^{\top}V-I_k> - <\Sigma, Z> ) ,\notag
\end{align}
where $H$ and $\Sigma$ are lagrangian multipliers, and $\Sigma \geq 0$.
The KKT conditions are given as:
\begin{subnumcases}{}
\nabla_Z L = \{-2\gamma(C-Z) -\Sigma \}|_{V=\widetilde{V}}=0; \label{eq:a}\\
\begin{aligned}
\nabla_C L =& \{2(C-A) + 2\alpha X^{\top}(X-XC) +2\beta(C-VV^{\top}) + 2\gamma (C-Z)\}|_{V=\widetilde{V}}=0; \label{eq:b}
\end{aligned}\\
\nabla_V L = \{-4(C-VV^{\top} )V + 2VH\} |_{V=\widetilde{V}} =0;  \label{eq:c}\\
 \widetilde{V}^{\top} \widetilde{V} = I_K;   \label{eq:d}\\
\Sigma \odot Z=0;  \label{eq:e}\\
\Sigma \geq 0,  \label{eq:f}
\end{subnumcases}
where $V=\widetilde{V}$ is an orthonormal matrix.
Therefore, combining Equation\eqref{eq:a} and Equation \eqref{eq:b}, $\Sigma = 2(C-A)+2\alpha X^{\top}(X-XC) + 2\beta(C-\widetilde{V}\widetilde{V}^{\top})$.
Combine Equation \eqref{eq:c} and 
Equation \eqref{eq:d}, we get : $C\widetilde{V} = \widetilde{V} (I_K+\frac{1}{2}H) = \widetilde{V} \widetilde{H}$,
where $\widetilde{H} = I_K + \frac{1}{2}H=\widetilde{V}^{\top}C\widetilde{V}$.
So $\widetilde{H}$ and $C$ have the same eigenvalues $\Lambda$. 
Thus, $\widetilde{H} = G \Lambda G^{\top}$, where $G$ is an orthonormal matrix. 
So $C \widetilde{V}G^{\top} = \widetilde{V}G^{\top}\Lambda$, which is  equivalent to $CV = V\Lambda$ with $V = \widetilde{V}G$, $V$ is also an orthonormal matrix.
Because of $\|C-VV^{\top}\|_F^2 = \|C-\widetilde{V}\widetilde{V}^{\top}\|_F^2$ and $V^{\top}V=I_K$, $V$ is also the minimum point.

Finally, reconsider the sparse constraint. \cite{SLSA} put forward that the sparse problem
\begin{equation}
    \min_Z \left\{F_t(Z,C, V): Z \geq 0, \|Z\|_0 \leq \eta\right\}
\end{equation}
is equivalent to
\begin{equation}
    \min_Z \left\{F_t(Z,C, V): Z \geq 0, \sup(Z)=\Omega_\eta\right\},
\end{equation}
where \(\Omega_\eta\) is the support set of the truncated sparse matrix \(T_\eta(C^+)\).
The latter gives solutions satisfying $Z=T_\eta(C^+)$.

Finally, we get our KKT conditions, solutions $Z,C,V$ and there exists a $\Sigma \geq 0$ satisfied:
\begin{subnumcases}{}
\begin{aligned}
 \Sigma =& 2[C-A+\alpha X^{\top}(X-XC) + \beta(C-{V}{V}^{\top})];   \label{eq:aa}
\end{aligned}\\
CV = V\Lambda;  \label{eq:bb}\\
\Sigma \odot Z=0;  \label{eq:cc}\\
Z=T_\eta(C^+).  \label{eq:dd}
\end{subnumcases}

\end{pf}

\begin{theorem}\label{thm:conv}
Let $\{Z_k, C_k, V_k\}$ be generated by the algorithm T-CML, and let $Q_k = V_kV_k^{\top}$. Then the following statements are true.
\begin{itemize}
\item[(a)] The sequence $\{F_t(Z_k, C_k, V_k)\}$ converges.
\item[(b)] If $\lambda_K(C_*)>\lambda_{K+1}(C_*)$ for any accumulation point $C_*$ of $\left\{C_k\right\}$, the matrix sequence $\{Q_k-Q_{k+1}\}$ converges to zero.
\item[(c)] If $\{Q_k\}$ has an isolated accumulation point $Q_*=V_*V_*^{\top}$ and $\lambda_K(C_*)>\lambda_{K+1}(C_*)$ for any accumulation point $C_*$ of $\left\{C_k\right\}$, $Q_k$ converges $Q_*$ and $\{C_k\}$ converges. Furthermore, if $C_*=\lim_{k\to\infty}C_k $ is $\eta$-separable, then $\{Z_k\}$ converges to $T_{\eta}(C_*^+)$.
\item[(d)] Any accumulation point $\left(Z_*,C_*,  V_*\right)$ of the sequence $\left\{\left(Z_k, C_k, V_{k}\right)\right\}$ satisfies the KKT conditions of T-CML if $\lambda_K\left(C_*\right)>\lambda_{K+1}\left(C_*\right)$.
\end{itemize}
\end{theorem}

\begin{pf}

Proof of (a). The alternating iterations over the three subproblems \eqref{eq:subV} \eqref{eq:subC} \eqref{eq:subZ}  make sure the monotonic decrease of the nonnegative sequence $\{F_t(Z_k, C_k, V_k)\}$:
\begin{align}
    F_t(Z_{k}, C_{k}, V_{k}) &\geq F_t(Z_{k}, C_{k}, V_{k+1})\geq F_t(Z_{k}, C_{k+1},V_{k+1}) \geq F_t(Z_{k+1}, C_{k+1},V_{k+1}), 
\label{eq:Finq}
\end{align}
and thus $F_t(Z_k, C_k, V_k)$ is convergent since it is lower bounded. 

Proof of (b). 
Assume on the contrary that $\{Q_k-Q_{k+1}\}$ does not converge to zero, which implies that $\exists$ an $\epsilon>0$ and a subsequence $\{Q_{k_j}-Q_{k_j+1}\}$ such that $\|Q_{k_j}-Q_{k_j+1}\|_F\geq \epsilon$ for any $j$. 
Without loss of generality, we can assume that $V_{k_j}\to \hat{V}_*$, $V_{k_j+1}\to V_*$ and $C_{k_j}\to C_*$, $Z_{k_j}\to Z_*=T_\eta(C_*^+)$ due to the boundedness of $Z_k$, $C_k$  and $V_k$. 
By taking limit on Equation \eqref{eq:Finq}, we can get $F_t(Z_*, C_*, V_*) = F_t(Z_*, C_*, \hat V_*)$. 
We further consider the limit on $C_{k_j}V_{k_j+1} = V_{k_j+1}\Lambda_{k_j+1}$, it yields $C_*V_* = V_*\Lambda_*$, where $\Lambda_*$ is a diagonal matrix of the $K$-largest eigenvalues of $C_*$, due to the continuously dependence of eigenvalues on the matrix entries. 
Thus, $V_*$ contains the eigenvectors of $C_*$ corresponding to the $K$ largest eigenvalues, which means that $V_*$ minimizes $\|C_*-VV^{\top}\|_F$ subject to $V^{\top}V=I_K$. 
$F_t(Z_*, C_*, \hat V_*) = F_t(Z_*, C_*, V_*)$ implies that $\hat V_*$ solves the same optimization problem, and thus $\hat V_*\hat V_*^{\top} = V_*V_*^{\top}$ since the solution is unique by $\lambda_K(C_*)>\lambda_{K+1}(C_*)$. 
Therefore, we could have
$Q_{k_j}-Q_{k_j+1} \to V_*V_*^{\top}-\hat{V}_*\hat{V}_*^{\top} = 0,$
which results in a contradiction to the assumption. 

Proof of (c). Since the condition in (b) holds, we have $\{Q_k-Q_{k+1}\}\to 0$.  Because $\{Q_k\}$ has an isolated accumulation point $Q_*$, by Lemma 4.10 of \cite{More1983}, we could conclude that if $Q_k$ does not converge, there exists a subsequence $\{Q_{k_j}\}$ of $\{Q_k\}$ and an $\epsilon>0$ such that $\|Q_{k_j+1}-Q_{k_{j}}\|_F\geq \epsilon$, which results in a contradiction with $Q_k-Q_{k+1}\to 0$. Therefore, $\{Q_k\}$ converges and converges to its accumulation point $Q_*$. On the other hand, since $F_t(Z,C, V)$ is continuous, the convergence of $\{Q_k\}$ implies that the sequence $\{C_k\}$ also converges. 
Since $C_*$ is separable, $C_k$ is also $\eta$ separable for sufficiently large $k$, and the index set of $\eta$-largest elements in $C_k$ does not change for sufficiently large $k$. Thus, $ \{Z_k=T_{\eta}(C_k^+)\}$ converges to $T_{\eta}(C_*^+)$.

Proof of (d).
Lemma 2 above shows that the KKT conditions of T-CML are that there exists a $\Sigma \geq 0$ such that
\begin{equation}
\begin{cases}       
\Sigma = 2[C-A+\alpha X^{\top}(X-XC) + \beta(C-{V}{V}^{\top})];\\ 
CV = V\Lambda;\\
\Sigma \odot Z=0;\\
Z=T_\eta(C^+).
\end{cases}\label{KKT}
\end{equation}
where $\Lambda=\text{diag}(\lambda_1(C),\cdots,\lambda_K(C))$ with the eigenvalues in decreasing order $\lambda_1(C) \geq \cdots\geq \lambda_K(C)).$

Taking one accumulation point $(Z_*,C_*,V_*)$ of sequence $\{(Z_k,C_k,V_{k+1})\}$, there exists a subsequence $\{(Z_{k_j},C_{k_j},V_{k_j+1)}\}$ which converges to $(Z_*,C_*,V_*)$. 
On the one hand, from $C_{k_j} V_{k_j+1}=V_{k_j+1} \Lambda_{k_j+1}$, we get 
$C_* V_*=V_* \Lambda_*$, where the diagonal matrix $\Lambda_*$ contains the $K$ eigenvalues of $C_*$. 

On the other hand, let's consider the subsequence $\{Z_{k_j}\}$, where $Z_{k_j}$ solves the optimization problem \eqref{eq:subZ}. $\min _{Z \geq 0} F_t(Z, C_{k_j}, V_{k_j})$ subject to $\|Z_{\text{off}}\|_0\leq \eta$. 
The solution $Z_{k_j}=T_\eta(C_{k_j}^+)$.
The convergence $\lim_{j\to\infty}Z_{k_j}=Z_*$ and the sparse constraint imply that  $Z_{k_j}$ for sufficiently large $j$ has the same support set $\Omega_\eta$, and thus $Z_*$ has the support set $\Omega_\eta$ too. 
This means that $Z_*=T_\eta(Z_*^+) = T_\eta(T_\eta(C_*^+)^+) = T_\eta(C_*^+)$. 

Furthermore, since ${Z}_{k_j}$ with support set $\Omega_\eta$ for sufficiently large $j$ satisfies the sparse constraint, due to the optimality it should satisfy the KKT conditions of \eqref{eq:subZ} for the support set $\Omega_\eta$, that is, there exists an nonnegative matrix $\Sigma_{k_j}$ such that
\begin{align}
&\Sigma_{k_j} = 2\gamma(Z_{k_j}-C_{k_j}); \label{eq:kkt_1}\\
&C_{k_j}-A+\alpha X^{\top}(X-X C_{k_j}) + \beta(C_{k_j}-Q_{k_j})=\gamma(Z_{k_j}-C_{k_j}) ; \notag\\
&\Sigma_{k_j} \odot Z_{k_j}=0.  \notag
\end{align}
Without loss of generality, we assume that $V_{k_j}\to \hat{V}_*$, similar to the proof of (b), and we can get $\{Q_{k_j}\}$ converges to $Q_*=V_*V_*^{\top}=\hat{V}_*\hat{V}_*^{\top}$.  Without loss of generality, $\{\Sigma_{k_j}\}$ is convergent due to the convergence of $\{Z_{k_j}\}$ and $\{C_{k_j}\}$ in \eqref{eq:kkt_1}. 
Taking the limit on the two sides of equations above for $j\to \infty$, we know that there exists a $\Sigma_* \geq 0$ such that 
\begin{equation*}
\begin{aligned}
&\Sigma_* = 2(C_*-A)+2\alpha X^{\top}(X-XC_*) + 2\beta(C_*-Q_*),\\
&\Sigma_* \odot Z_*=0.\\
\end{aligned}
\end{equation*}
Overall, $(Z_*,C_*,V_*)$ satisfies the KKT conditions in \eqref{KKT}, and the proof is completed.
\end{pf}
\vspace{0.5cm}

To empirically validate the convergence behavior predicted by the theoretical analysis, we track the clustering performance throughout the iterative optimization process on four real-world datasets (Cora \cite{cora}, Citeseer, DBLP, and Pubmed). Specifically, after the $k$-th iteration of the CML algorithm, Spectral Clustering is applied to the learned graph $Z_k$, and the corresponding ACC, NMI, and F1-score are recorded. As illustrated in Figure~\ref{fig:convergence}, the performance metrics on Cora, Citeseer, and DBLP improve rapidly during the early iterations and become stable after approximately 5--10 iterations. For the large-scale Pubmed dataset, we report the results of Fast-CML, where ACC, NMI, and F1-score similarly converge within 10 iterations. These observations are consistent with the theoretical convergence analysis and demonstrate that both CML and Fast-CML achieve stable solutions with only a small number of iterations in practice.

\begin{figure*}[t!]
\centering
\begin{minipage}{0.24\linewidth}
    \centering
    \includegraphics[width=\linewidth]{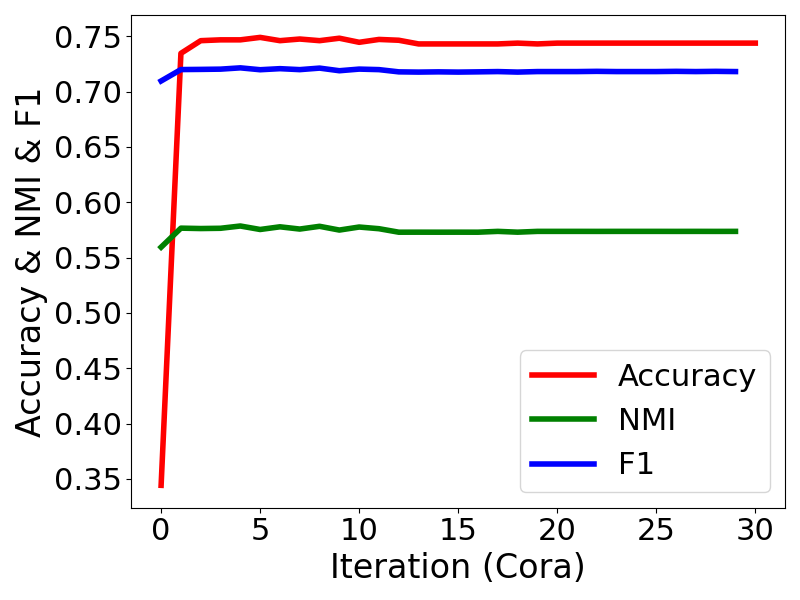}
\end{minipage}\hfill
\begin{minipage}{0.24\linewidth}
    \centering
    \includegraphics[width=\linewidth]{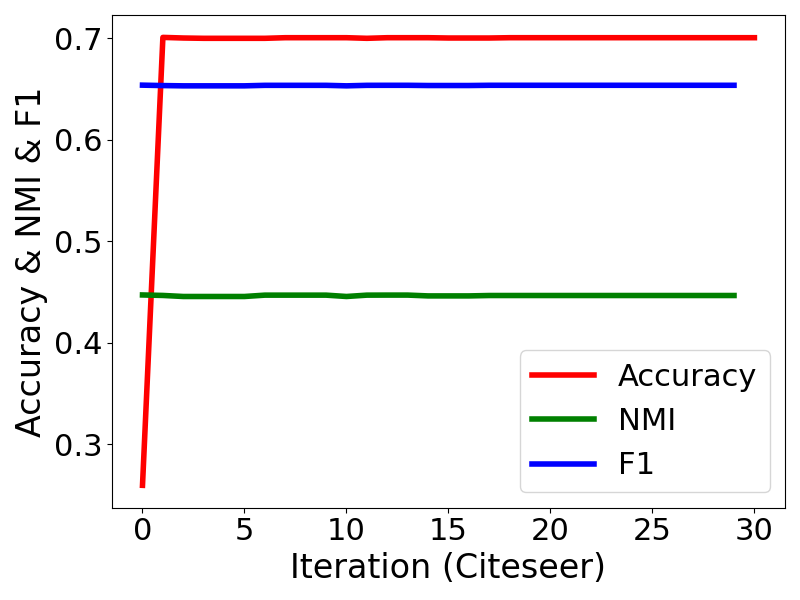}
\end{minipage}\hfill
\begin{minipage}{0.24\linewidth}
    \centering
    \includegraphics[width=\linewidth]{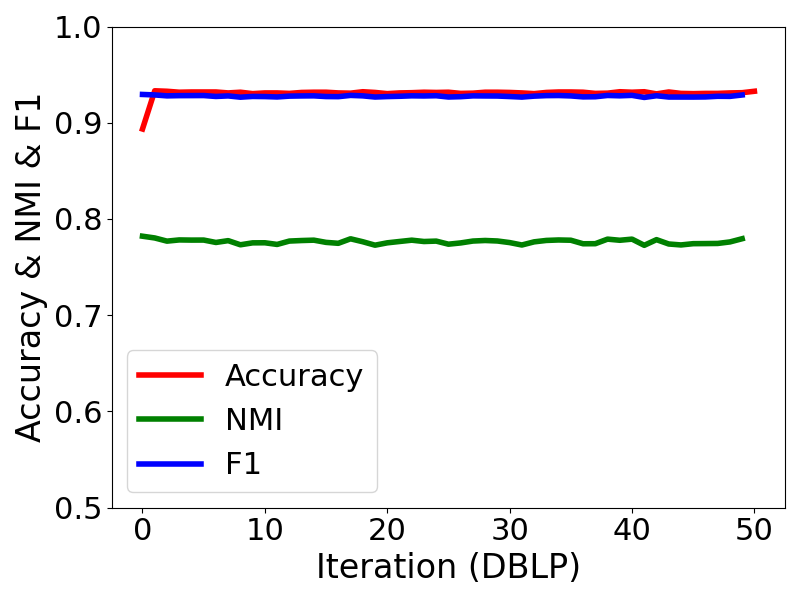}
\end{minipage}\hfill
\begin{minipage}{0.24\linewidth}
    \centering
    \includegraphics[width=\linewidth]{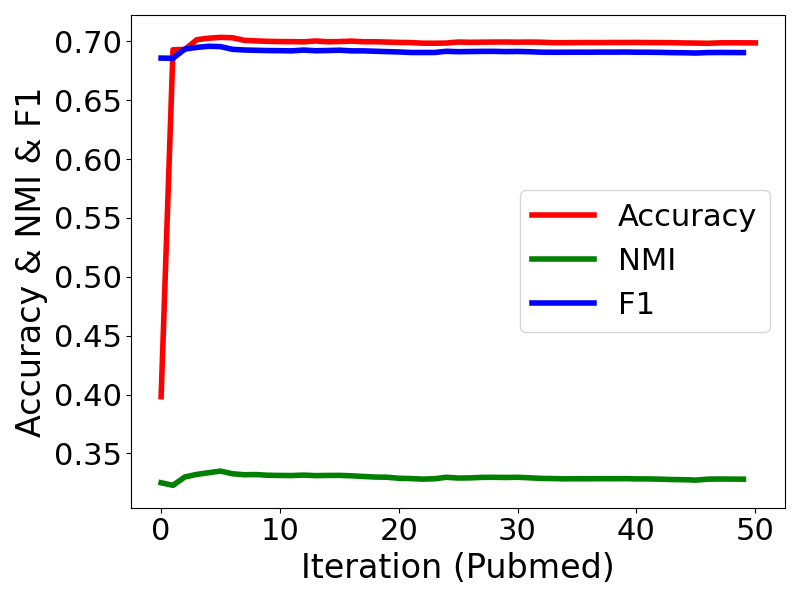}
\end{minipage}
\caption{Visualization of convergence effect on Cora, Citeseer, DBLP and Pubmed, where the Cora, Citeseer and DBLP dataset use the CML algorithm, and Pubmed uses the Fast-CML algorithm.}
\label{fig:convergence}
\end{figure*}

\section{Experiments}

In this section, we first present a comprehensive experimental evaluation of the denoising capability of the proposed CML framework on simulated attributed graphs. Through simulations spanning varied noise types and intensity levels, we systematically benchmark A-CML and T-CML against baseline methods, under different scenarios. 
We then evaluate the CML framework on real-world datasets for attributed graph clustering and sparse subspace learning. 
The source code is available at \url{https://github.com/LiminLi-xjtu/CML_model}.

\subsection{Datasets and Experimental Settings}
\subsubsection{Simulated Data}
The synthetic attributed graph is constructed with three distinct classes of nodes, where the node attributes are approximately distributed around three subspaces embedded in a $d$-dimensional ambient space.

Let $\boldsymbol{\mu}_1, \boldsymbol{\mu}_2, \boldsymbol{\mu}_3 \in \mathbb{R}^d$ be orthonormal vectors, each $\boldsymbol{\mu}_k$ can span a one-dimensional subspace. For example, given a vector $a_k$ of $n_k$ nonzero components, $\boldsymbol{\mu}_k a_k^\top$ gives $n_k$ samples from this subspace. We construct $n_k$ noisy samples 
\begin{equation*}
    {X}_k(\epsilon) 
    = \boldsymbol{\mu}_k a_k^\top 
    + \epsilon^2 \boldsymbol{\mu}_k^{\bot} {B}_k,
\end{equation*}
where $\epsilon$ is a noise level, $\boldsymbol{\mu}_k^{\bot}\in \mathbb{R}^{d\times (d-1)}$ is
an orthonormal basis of the orthogonal complement space of $\boldsymbol{\mu}_k$, and $B_k \in \mathbb{R}^{(d-1) \times n_k}$ is independently sampled with $\mathcal{N}(0,1)$. Then, the attribute matrix is given by
\begin{equation*}
    {X}(\epsilon) = [{X}_1(\epsilon), {X}_2(\epsilon), {X}_3(\epsilon)] \in \mathbb{R}^{d \times N}.
\end{equation*}
In the experiment, we choose 
\begin{equation*}
    \boldsymbol{\mu}_1 = \frac{[1,1,0,0,0]^\top}{\sqrt{2}},
    \boldsymbol{\mu}_2 = [0,0,1,0,0]^\top,
    \boldsymbol{\mu}_3 = \frac{[0,0,0,1,1]^\top}{\sqrt{2}}.
\end{equation*}
set $(n_1, n_2, n_3) = (50, 80, 100)$ and $N = n_1+n_2+n_3$.

The noisy adjacency matrix is constructed as
\[
    A(\delta) = A_0 + \bar{A}(\delta),
\]
where $A_0\in \{0,1\}^{N \times N}$ is a block-diagonal matrix with three diagonal blocks in sizes $n_1, n_2, n_3$, respectively, each has entries sampled as
$    ({A}_0)_{ij} \sim \text{Bernoulli}(p)
$
with $p = 0.05$, and encodes intra-class connectivity.
$\bar{A}(\delta)$ is a binary matrix. 
Its nonzero entries appear only outside the diagonal blocks of $A_0$, 
indicating the edges between different classes. We choose the number of these edges as
$\|\bar{{A}}(\delta)\|_0 = \delta^2 \|{A}_0\|_0.$

The parameter $\epsilon \in [0,1]$ controls the noise of attribute, while $\delta \in [0,1]$ regulates the confusion of graph topology.
As $\delta$ increases, distinguish between classes
is progressively degraded, with a substantial increase in inter-group connections, as illustrated in Figure \ref{fig:simulation_data} (c-f).
For the attribute perturbation, increasing $\epsilon$ close to 1 will
significantly distort the intrinsic subspace structure. 
Figure \ref{fig:simulation_data} (g–l) presents the visualization of $X(\epsilon)$ under progressively increasing values of $\epsilon$.

\begin{figure*}[t!]
\centering

\begin{minipage}{0.16\linewidth}
    \centering
    \includegraphics[width=\linewidth]{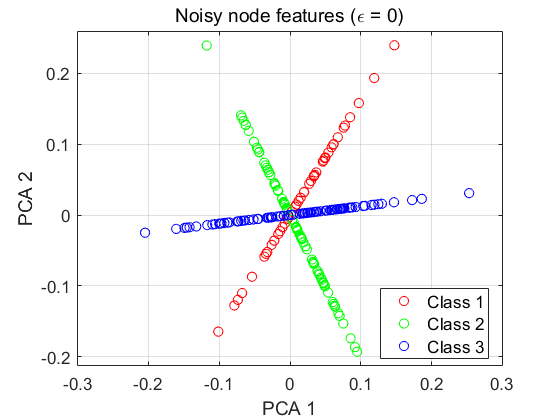}\\[2ex]
    \scriptsize (a) $X_0$
\end{minipage}\hfill
\begin{minipage}{0.16\linewidth}
    \centering
    \includegraphics[width=\linewidth]{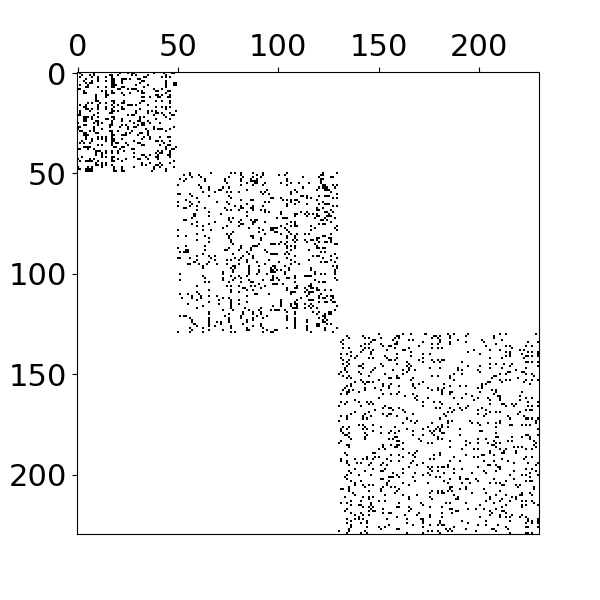}\\ 
    \scriptsize (b) $A_0$
\end{minipage}\hfill
\begin{minipage}{0.16\linewidth}
    \centering
    \includegraphics[width=\linewidth]{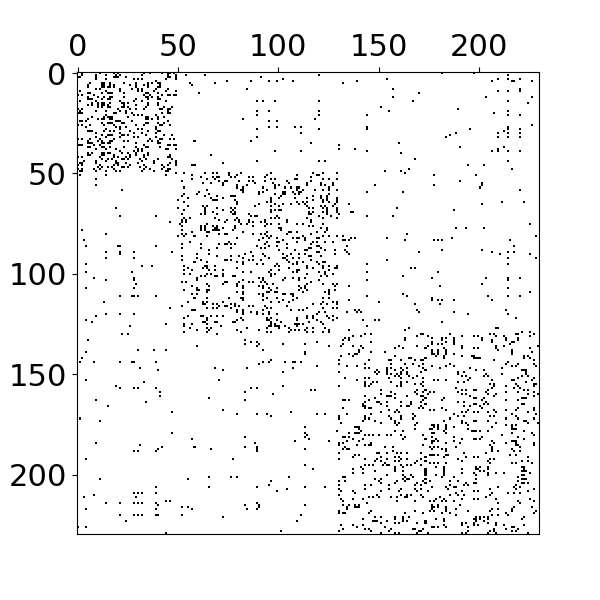}\\ 
    \scriptsize (c) $A(\delta=0.1)$
\end{minipage}\hfill
\begin{minipage}{0.16\linewidth}
    \centering
    \includegraphics[width=\linewidth]{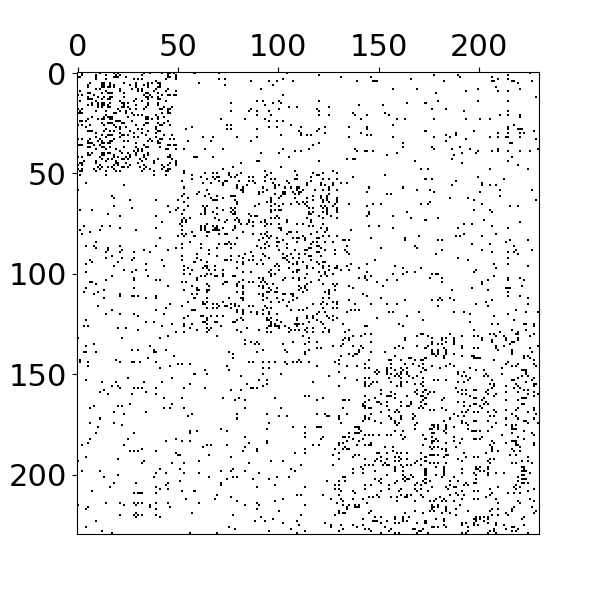}\\ 
    \scriptsize (d) $A(\delta=0.3)$
\end{minipage}
\hfill
\begin{minipage}{0.16\linewidth}
    \centering
    \includegraphics[width=\linewidth]{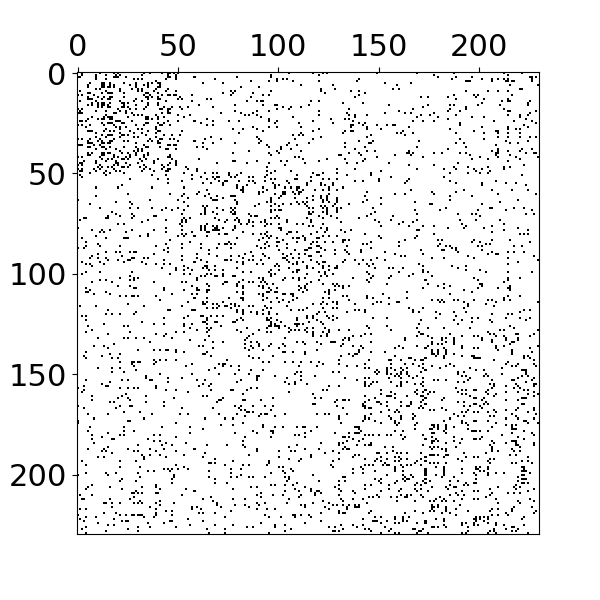}\\ 
    \scriptsize (e) $A(\delta=0.6)$
\end{minipage}\hfill
\begin{minipage}{0.16\linewidth}
    \centering
    \includegraphics[width=\linewidth]{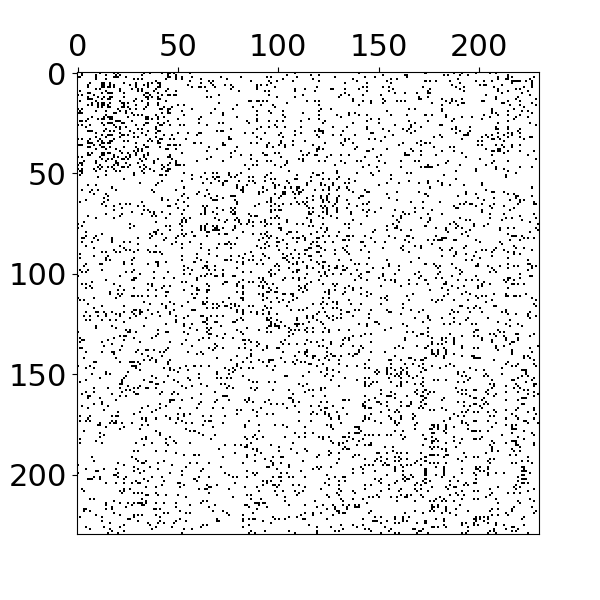}\\ 
    \scriptsize (f) $A(\delta=0.9)$
\end{minipage}

\vspace{0.5em}

\begin{minipage}{0.16\linewidth}
    \centering
    \includegraphics[width=\linewidth]{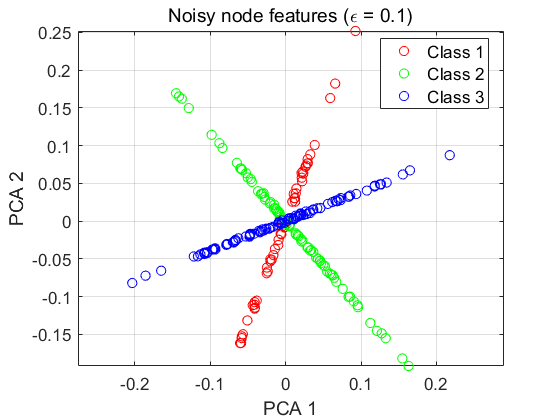}\\ 
    \scriptsize (g) $X(\epsilon=0.1)$
\end{minipage}\hfill
\begin{minipage}{0.16\linewidth}
    \centering
    \includegraphics[width=\linewidth]{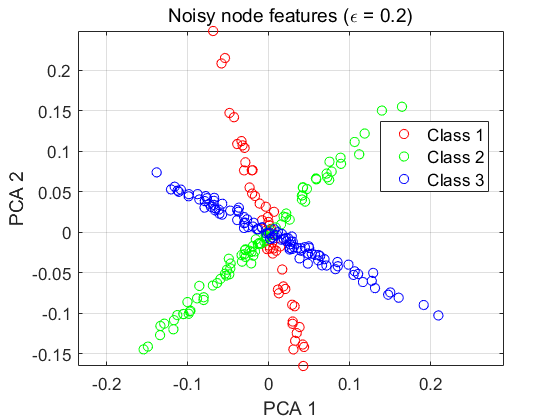}\\ 
    \scriptsize (h) $X(\epsilon=0.2)$
\end{minipage}
\hfill
\begin{minipage}{0.16\linewidth}
    \centering
    \includegraphics[width=\linewidth]{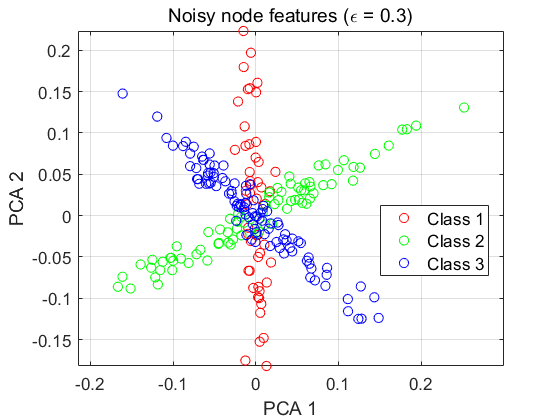}\\ 
    \scriptsize (i) $X(\epsilon=0.3)$
\end{minipage}\hfill
\begin{minipage}{0.16\linewidth}
    \centering
    \includegraphics[width=\linewidth]{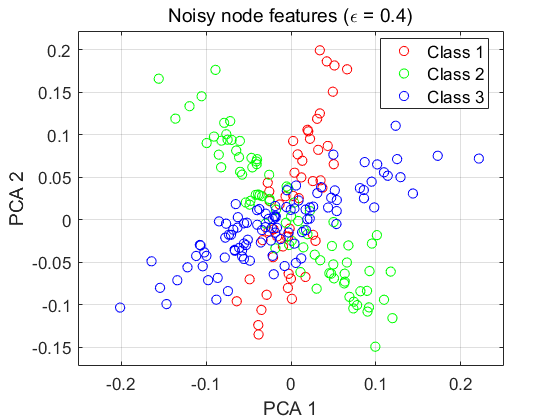}\\ 
    \scriptsize (j) $X(\epsilon=0.4)$
\end{minipage}\hfill
\begin{minipage}{0.16\linewidth}
    \centering
    \includegraphics[width=\linewidth]{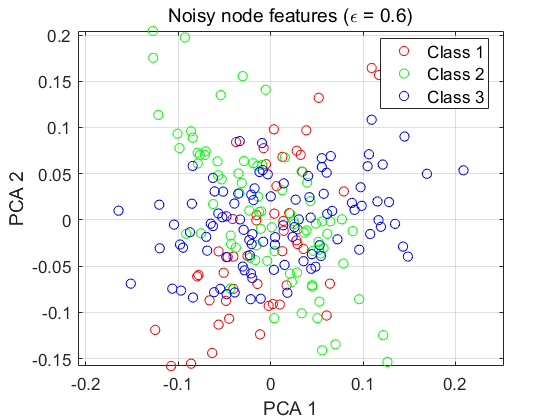}\\ 
    \scriptsize (k) $X(\epsilon=0.6)$
\end{minipage}\hfill
\begin{minipage}{0.16\linewidth}
    \centering
    \includegraphics[width=\linewidth]{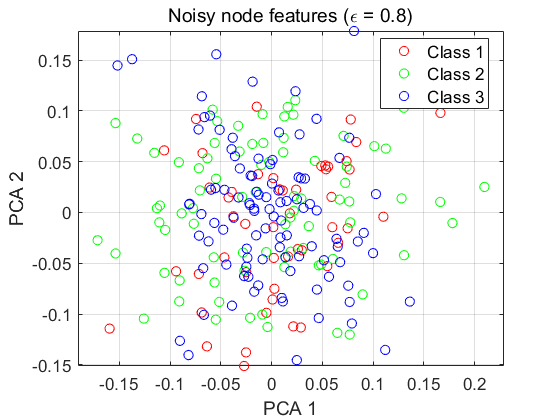}\\ 
    \scriptsize (l) $X(\epsilon=0.8)$
\end{minipage}

\caption{Visualization of the simulation datasets $(A(\delta), X(\epsilon))$.}
\label{fig:simulation_data}
\end{figure*}

\subsubsection{Real-world Datasets}\label{sect:t}

Six  attributed graph datasets are used to compare the efficiency of the compared algorithms on attributed graph node clustering. The summary of these datasets is listed in Table \ref{tab:data}.
\begin{table}[t]
\caption{Summary of the attributed graph datasets used in the experiments.}
\label{tab:data}
\centering
\fontsize{10}{12}\selectfont
\begin{tabular}{@{}lccccc@{}}
\toprule
Dataset & Nodes ($n$) & Edges & Features ($d$) & Classes ($K$) & $\Delta h$ \\
\midrule
Cora     & 2708  & 5429    & 1433 & 7  & 0.1824 \\
Citeseer & 3327  & 4732    & 3703 & 6  & 0.0703 \\
Wiki     & 2405  & 17981   & 4973 & 17 & 0.0028 \\
ACM      & 3025  & 16153   & 1870 & 3  & 0.0978 \\
DBLP     & 4057  & 2502276 & 334  & 4  & 0.3095 \\
Pubmed   & 19717 & 44338   & 500  & 3  & 0.0654 \\
\bottomrule
\end{tabular}
\end{table}

\begin{itemize}
    \item Cora~\cite{cora}: Literature citation network. A directed edge exists between two papers if one cites the other. Each paper is also associated with a 1433-dimensional bag-of-words model that represents its lexical features. The papers are categorized into seven subject areas including theory, reinforcement learning, genetic algorithms, and probabilistic methods and so on.
    \item Citeseer~\cite{cora}: Literature citation network. It comprises 3327 nodes, each representing an academic paper. The edges in the graph indicate citation relationships between papers. Similar to the Cora dataset, each paper in Citeseer is associated with a bag-of-words model that represents the lexical characteristics of the paper. The dataset includes six subject categories, namely AI, ML, DB, HCI, and others.
    \item Wiki~\cite{wiki}: Web pages network. The graph consists of web pages that are connected through links. The Wiki contains 2405 web pages and 17981 link relations. The nodes are categorized into 17 classes and have 4973-dimensional vectors as attributes.
    \item ACM~\cite{acm}: Literature citation network. Similar to Cora and Citeseer. There are 3025 articles of 3 categories: Database, Wireless Communication and Data Mining.
    The feature dimension of each paper is 1870, comprising the keywords that are specific to that paper. Each keyword is assigned a position in the bag of words, with a value of 1 indicating that the keyword appears in the paper and 0 indicating that it does not.
    \item  DBLP~\footnote{https://dblp.uni-trier.de}: Authors network. 
    In DBLP dataset, there are 4057 nodes, each representing an author. An edge between two authors indicates co-authorship. Author features are elements in a bag of words represented by keywords. These authors are grouped into 4 research areas (databases, data mining, machine learning and information retrieval).
    \item Pubmed~\cite{cora}: Literature citation network. The Pubmed database contains a collection of 19,717 papers on diabetes. Each paper characterized by a 500-dimensional vector. 
    The edges indicate citation relationships between papers.
    The papers are divided into three categories: Experimental Diabetes Mellitus, Type 1 Diabetes Mellitus, Type 2 Diabetes Mellitus. 
\end{itemize}

For each dataset, a low-pass graph filter $G=(I-L)^t$ defined in \cite{AGC} is used 
to filter the attribute feature $X$ to $X:=GX$ so that the adjacent nodes have similar features. Here, $L$ is the Laplacian matrix $D^{-1/2}AD^{-1/2}$, and the smoothing parameter $t$ is used to control the degree of 
smoothness of the graph.

We conduct experiments using three benchmark single-input datasets, which contain only features and category information for the samples, without any graphs or edges.
\begin{itemize}
    \item Iris \cite{iris}\footnote{\url{https://scikit-learn.org/stable/modules/generated/sklearn.datasets.load_iris.html}}: The dataset consists of 150 samples divided into three classes, with each sample has 4 features: sepal length, sepal width, petal length, and petal width.
    These features were observed by botanist Edgar Anderson in Irish Iris spp.
    \item Yale\footnote{\url{http://cvc.cs.yale.edu/cvc/projects/yalefaces/yalefaces.html}}: 
    Yale dataset is a classic dataset used for face recognition research. It contains 165 face images from 15 different people with 11 images for each person. 
    Each image represents a different emotion. 
    \item Digits
    \footnote{\url{https://scikit-learn.org/stable/modules/generated/sklearn.datasets.load_digits.html}} :
    The Digits dataset comprises $8\times 8$ handwritten digital images with 1797 samples. Each sample is a 64-dimensional feature vector, representing the pixel intensities of the image. The dataset contains 10 categories that are frequently employed in machine learning classification tasks.
\end{itemize}

\subsubsection{Modal Quality} 
We evaluate the modal quality by measuring consistency of the neighbor labels through graph and feature homophily.
The homophily of the edges of graph $A = (a_{ij})$, denoted by $h_A \in [0, 1]$, is the proportion of intra‑class edges:
\[
    h_{{A}} = \frac{\sum_{i<j} a_{ij} \, \mathbbm{1}[y_i = y_j]}{\sum_{i<j} a_{ij}},
\]
where $\mathbbm{1}[\cdot]$ denotes the indicator function, 
and $\{y_i\}$ are the true labels. For attributes, we use the homophily of a graph of $k$-nearest‑neighbors of the columns of $X$ to define the homophily of the attributes, denoted as $h_X$. 
The neighborhood size $k\geq 2$ is set to equal the average number $\bar d$ of connected neighbors in $A$. That is, $k = \max(2,\ \operatorname{round}(\bar{d}))$.

The quality gap between the two modalities is quantified by the homophily difference $\Delta h = h_{A} - h_{X}.$
A positive value of $\Delta h$ indicates that the topological modality exhibits stronger label homophily than the attribute modality and is therefore less affected by noise. 

\subsubsection{Baselines for Attributed Graph Learning}
We used 15 well-known and widely adopted methods as baselines, including 12 methods designed for attributed graphs and 7 based on deep learning.
\begin{itemize}
    \item K-means: A classical clustering algorithm that divides data into $K$ clusters by iteratively assigning each point to the cluster whose mean minimizes the sum of the intra-cluster squared distances is presented.
    The algorithm uses an iterative optimization approach for compact cluster representation in Euclidean space.
    \item Spectral Clustering: Unsupervised learning often uses the eigenstructure of affinity matrices to partition data into clusters. This method is powerful because it captures complex geometric structures and nonlinear relationships in high-dimensional spaces.
    \item Deepwalk: A graph learning technique that uses random walks and the SkipGram model to generate node embeddings, capturing the local structures of the graph in a continuous vector space.
    \item GAE,VGAE,MGAE~\cite{VGAE,MGAE}: Graph Autoencoder (GAE), Variational Graph Autoencoders (VGAE), and Marginalized Graph Autoencoder (MGAE) are a range of graph-based deep learning models that aim to learn latent representations of nodes.  While GAE uses adversarial training to generate realistic embeddings, VGAE introduces a probabilistic framework with variational inference for uncertainty modeling, and MGAE extends the versatility by accommodating mixed-type edges and capturing both deterministic and stochastic aspects in graph-structured data.
    \item AGC~\cite{AGC}: Adaptively incorporates node-specific attributes into the graph convolution process, enabling nuanced exploration of attribute-rich structures for more refined and accurate graph clustering results.
    \item DAEGC~\cite{DAEGC}: Distinguished by its use of deep attentional embeddings, this method adeptly captures intricate attribute interactions within the graph structure. This improves the clustering process with a nuanced focus on the importance of node attributes.
    \item GIC~\cite{GIC}: An approach that enhances graph representation deep learning by maximizing the mutual information between the graph’s substructures at a coarse-grained level.
    \item SG$^2$C~\cite{SSGC}: The spectral graph convolution operation is used, with a focus on simplicity and interpretability, while ensuring effective feature propagation over graph-structured data. This makes it a robust and transparent solution for graph-based deep learning tasks.
    \item DFCN~\cite{DFCN}: It achieves significant performance improvements on deep clustering tasks by combining self-encoders and graph neural networks and introducing a fusion module for structural and attribute information as well as a triplet self-supervised strategy.
    \item FGC~\cite{FGC}: Fine-grained information is utilized to enhance clustering precision and adaptability, making it a potent technique for nuanced analysis and interpretation of complex graph-structured data.
    \item GCC~\cite{GCC}: Seamlessly integrates node representation learning and clustering, demonstrating a remarkable ability to jointly optimize both tasks with computational efficiency.
    This makes it a powerful approach for comprehensive and resource-efficient analysis of graph-structured data.
    \item SAGSC~\cite{SAGSC}: Proficient in using kernel methods to efficiently uncover the intricate complex underlying substructures in large attributed graphs, demonstrating superior scalability and performance.
    \item DCGC~\cite{DCGC}: Integrates graph structure and node attributes by minimizing their joint normalized cut and employs optimal transport for self-supervised clustering to circumvent representation collapse and degenerate solutions.
\end{itemize}

\subsubsection{Baselines for Single-Input Clustering }

We compare our method with the following two single-input baseline models.
\begin{itemize}
    \item SSC~\cite{SSC}: Sparse Subspace Clustering (SSC) utilises the sparsity-inducing property of data representations to accurately identify subspace structures, making it a robust and effective method for clustering high-dimensional data with shared subspaces.
    \item SSCE~\cite{SSCE}: The Sparse Subspace Clustering with Entropy-Norm (SSCE) algorithm introduces an entropy-norm regularizer to enhance subspace clustering by promoting both sparsity and entropy in the data representation.
    This leads to improved performance in capturing underlying subspace structures, particularly in the presence of noise and outliers.
\end{itemize}

\subsubsection{Evaluation Metrics}

We adopt three widely used clustering evaluation metrics, i.e., clustering accuracy (ACC), normalized mutual information (NMI), and F1-score, to comprehensively assess performance.

ACC measures the best one-to-one alignment between predicted cluster labels and ground-truth labels:
\begin{equation}
\mathrm{ACC} = \max_{\pi} \frac{1}{n} \sum_{i=1}^{n} \mathbb{I}\big(y_i = \pi(c_i)\big),
\end{equation}
where $c_i$ and $y_i$ denote the predicted and ground-truth labels of sample $i$, respectively, $\pi$ is a permutation mapping between clusters and ground-truth classes, and $\mathbb{I}(\cdot)$ is the indicator function.
NMI measures the mutual dependence between predicted clustering $C$ and ground-truth labels $Y$:
\begin{equation}
\mathrm{NMI} = \frac{2 I(C;Y)}{H(C) + H(Y)},
\end{equation}
where $I(C;Y)$ is the mutual information and $H(\cdot)$ denotes entropy. NMI quantifies the consistency of the clustering structure without requiring explicit label matching.
F1-score is computed based on pairwise precision and recall:
\begin{equation}
\mathrm{F1} = \frac{2 \cdot \mathrm{Precision} \cdot \mathrm{Recall}}{\mathrm{Precision} + \mathrm{Recall}},
\end{equation}
where precision and recall are defined on pairwise sample agreements:
\begin{equation}
\mathrm{Precision} = \frac{TP}{TP + FP}, \quad
\mathrm{Recall} = \frac{TP}{TP + FN}.
\end{equation}

The three evaluation metrics ACC, NMI, and F1-score characterize clustering performance from distinct perspectives, leading to potential numerical divergence. ACC quantifies global label agreement via optimal one-to-one alignment, focusing on classification accuracy but overlooking internal cluster topology. In contrast, NMI employs an information-theoretic approach to assess the statistical consistency between partitions, reflecting how well the clustering captures the underlying data organization. F1-score provides a pairwise perspective, measuring local structural consistency by evaluating whether sample pairs are correctly grouped or separated. These divergent focuses explain their varying magnitudes: while a high ACC ensures correct label assignment, it does not guarantee compact or well-organized cluster structures. Consequently, fragmented clusters may yield a high ACC but a moderate NMI, whereas F1 remains uniquely sensitive to pairwise grouping fidelity.

\subsubsection{Parameter Settings}

The learned graph matrix $Z$ can be used in a new implementation of CML as the graph $A$, while consistently maintaining the original node attributes $X$. Repeating this procedure leads to a multi-stage CML. If CML is implemented for $s$ times in this way, we denote this procedure as CML$^s$.

The matrix $Z$ is expected to be block diagonal, assuming the nodes are properly ordered, the sparsity parameter can be calculated as $\eta=({n}/{K})^2 \times K= {n^2}/{K}$
for class-balanced datasets.
Empirically, we set $\eta = 10^4 \lfloor 10^{-5}n^2 / \rho K \rfloor$, where $\rho \in (0, 1)$ and $\lfloor \cdot \rfloor$ represents the floor function. In most cases, $\rho$ is chosen within the range of 0.1 to 0.6, adjusted based on the specific dataset. In our experiments, we set $\eta$ approximately as follows: $\eta = 30 \times 10^4$ for Cora, $\eta = 60 \times 10^4$ for Citeseer, $\eta = 250 \times 10^4$ for ACM, and $\eta = 60 \times 10^4$ for Wiki, 
and $\gamma$ was fixed at 0.001.

The learned graph matrix $Z$ by CML can serve for the node attribute clustering, by applying the spectral clustering method on $Z$. That is, compute the $K$ dominant eigenvectors of $Z$ first, which yields a lower-dimensional projection $\{y_i\}$ of the node attributes, and then take $k$-means 
to cluster the node attributes. 
In the experiments, the reported clustering results are obtained by performing spectral clustering on the graph learned by CML.

All experiments are conducted on a workstation running with an i7-10700 CPU @ 2.90 GHz (8 physical cores, 16 logical cores) and 16 GB RAM. No GPU was used. The implementation was carried out in Python 3.8.19 with key libraries: NumPy 1.22.4, SciPy 1.6.2, and scikit-learn 0.24.1.

\subsection{Clustering  Performance
on Simulated Data}

We first examine the robustness of the methods AGC~\cite{AGC}, FGC~\cite{FGC}, SLSA(t), A-CML, and T-CML that integrate both information sources across various noise levels on the simulated data. Varying structural noise levels $\delta$ and attribute noise levels $\epsilon$ are tested as 
    $\delta \in \{0.1, 0.3, 0.5, 0.7, 0.9\}, 
    \epsilon \in \{0.3, 0.5, 0.7\}$.
The results, evaluated by three commonly used metrics ACC, NMI, and F1-score values are summarized in Table~\ref{tab:simulate}, where the best results are in \textbf{bold}, and the second-best are \underline{underlined}.

For moderate to high noise levels, neither modality using $A$ or $X$ alone can reliably recover the cluster structure. For the attribute-only evaluation using $X$, we implement K-means, K-subspace \cite{ho2003clustering}, Spectral Clustering, SSC \cite{SSC}, SSCE \cite{SSCE}, SSC($\mathcal{L}_1$), and SLSA($\mathcal{L}_1$).
SSC($\mathcal{L}_1$) refers to the SSC model with the addition of the $\mathcal{L}_1$-norm.
SLSA($\mathcal{L}_1$) indicates that SLSA uses the similarity matrix constructed by SSC ($\mathcal{L}_1$) as the input.

For the structure-only evaluation using $A$, we apply Spectral Clustering, the Louvain algorithm \cite{Louvain}, DeepWalk \cite{{DeepWalk}}, and SLSA. Tables \ref{tab:X_only} and \ref{tab:A_only} report the ACC, NMI, and F1-score values for experiments using only $X$ and only $A$, respectively. 
As illustrated in Tables \ref{tab:X_only} and \ref{tab:A_only}, when the noise levels are minimal (e.g., $\epsilon, \delta = 0.1$), K-subspace and SLSA($\mathcal{L}_1$) effectively recover the underlying class structure using node attributes ($X$) alone. Similarly, Spectral Clustering, DeepWalk, and SLSA demonstrate good performance in identifying clusters based solely on the graph structure ($A$). 
As the noise levels $\epsilon$ and $\delta$ increase, the underlying class structure becomes increasingly obscured, rendering the independent use of $A$ or $X$ insufficient for accurate cluster recovery. 

\begin{table}[ht]
\centering
\caption{Clustering clustering using only attribute $X(\epsilon)$.}
\fontsize{9}{10}\selectfont
\begin{tabular}{cc|ccccc}
\toprule
Metric& Method & $\epsilon=0.1$ & $0.3$ & $0.5$ & $0.7$ & $0.9$ \\
\midrule
\multirow{6}{*}{ACC (\%)}&
K-means      &52.61  &49.65  & 43.30 &41.52  &{39.74}\\
&K-subspace   &\bf{99.57}  &\bf{93.04}  & \bf{80.43} &\bf{63.48}  &{38.26}\\
&Spectral     &{62.52}  &{56.39}  &46.08  &42.04  &38.09\\
&SSC          &60.43  &46.96  &{46.52}  &{43.04}  &36.96\\
&SSCE         &50.00  &50.87  &46.09  &{43.04}  &{38.26}\\
& SSC($\mathcal{L}_1$) &95.65 &\underline{92.61} &73.48 &49.57 &\underline{41.74}\\
& SLSA($\mathcal{L}_1$)  &\underline{97.39}  &\underline{92.61}  &\underline{75.65}  &\underline{61.74}  &\bf{44.35}  \\
\midrule
\multirow{6}{*}{NMI (\%)}&
K-means      &32.77  &21.38  &7.58  &3.47  &1.12\\
&K-subspace   &\bf{97.75}  &\underline{72.43}  &\bf{43.34}  &\bf{18.78}  &0.97\\
&Spectral     &{40.68}  &23.21  &8.86  &3.74  &0.83\\
&SSC          &20.48  &11.53  &6.86  &2.21 &1.77 \\
&SSCE         &30.02  &{31.01}  &{11.47}  &3.13  &\underline{2.57}\\
& SSC($\mathcal{L}_1$) &84.97 &\bf{72.63} &36.37 &7.92 &1.60\\
& SLSA($\mathcal{L}_1$)  &\underline{88.16}  &\bf{72.63}  &\underline{37.26}  &\underline{16.08}  &\bf{2.99}  \\
\midrule
\multirow{6}{*}{F1 (\%)}&
K-means      &52.97  &49.21  &42.58  &41.37  &\underline{39.14}\\
&K-subspace   &\bf{99.62}  &\bf{92.64}  &\bf{79.58}  &\bf{62.69}  &37.00\\
&Spectral     &63.59  &56.58  &46.02  &41.14  &37.37\\
&SSC          &59.49  &44.33  &46.44  &42.27  &37.07\\
&SSCE         &50.89  &51.98  &46.59  &43.00  &38.42\\
& SSC($\mathcal{L}_1$) &94.89 &\underline{92.36}  &72.84 &49.00 &39.04  \\
& SLSA($\mathcal{L}_1$)  &\underline{97.20}  &\underline{92.36}  &\underline{75.03}  &\underline{61.06}  &\bf{41.69}  \\
\bottomrule
\end{tabular}
\label{tab:X_only}
\end{table}

\begin{table}[ht]
\centering
\caption{Clustering results using only structure $A(\delta)$.}
\fontsize{9}{10}\selectfont
\begin{tabular}{cc|ccccc}
\toprule
Metric& Method & $\delta$=0.1 & $\delta$=0.3 & $\delta$=0.5 & $\delta$=0.7 & $\delta$=0.9 \\
\midrule
\multirow{4}{*}{ACC (\%)}&
Spectral     & \underline{98.95} & 76.17 & 64.05 & \underline{57.26} &40.30\\
&Louvain &76.26  &62.61  &39.39  &25.91  &22.27  \\
&DeepWalk &98.70  &\bf{91.78}  &\underline{86.13}  &56.83  &\underline{44.13}  \\
&SLSA         & \bf{99.13} & \underline{83.48} & \bf{89.57} & \bf{58.26} &\bf{45.22}\\
\midrule
\multirow{4}{*}{NMI (\%)}&
Spectral     &\underline{94.92}  &49.77  &23.11  &15.48  &4.38\\
&Louvain &79.42  &52.04  &30.18  &15.85  &\underline{6.88}\\
&DeepWalk &93.89  &\bf{69.43}  &\underline{56.64}  &\bf{17.53}  &3.93  \\
&SLSA          &\bf{95.65} &\underline{53.47} &\bf{62.48} &\underline{17.52}  &\bf{6.99}\\
\midrule 
\multirow{4}{*}{F1 (\%)}&
Spectral     &\underline{98.98}  &58.64  &55.14  &\underline{48.72}  &\underline{39.48}\\
&Louvain &65.62  &56.65  &41.46  &27.93  &23.12  \\
&DeepWalk &98.75  &\bf{91.08}  &\underline{85.05}  &47.07  &38.90  \\
&SLSA          &\bf{99.09}  &\underline{82.36}  &\bf{88.84}  &\bf{51.64}  &\bf{42.64}\\
\bottomrule
\end{tabular}
\label{tab:A_only}
\end{table}

\begin{table}[t]
\caption{Clustering results on the simulated data $(A(\delta),X(\epsilon))$ with
$\delta\in\{0.1,0.3,0.5,0.7,0.9\}$ and $\epsilon\in\{0.3,0.5,0.7\}$.
The best and second-best results per column are shown in bold and
underlined, respectively.}
\label{tab:simulate}
\centering
\fontsize{9}{10}\selectfont
\setlength{\tabcolsep}{3pt}
\begin{tabular}{@{}l*{15}{c}@{}}
\toprule
{}
  & \multicolumn{5}{c}{$\epsilon=0.3$}
  & \multicolumn{5}{c}{$\epsilon=0.5$}
  & \multicolumn{5}{c}{$\epsilon=0.7$}\\
\cmidrule(lr){2-6}\cmidrule(lr){7-11}\cmidrule(lr){12-16}
$\delta$& $0.1$ & $0.3$ & $0.5$ & $0.7$ & $0.9$
  & $0.1$ & $0.3$ & $0.5$ & $0.7$ & $0.9$
  & $0.1$ & $0.3$ & $0.5$ & $0.7$ & $0.9$\\
\midrule
$\Delta h$
  & 0.09 & 0.02 & $-0.10$ & $-0.23$ & $-0.35$
  & 0.27 & 0.19 & 0.08 & $-0.05$ & $-0.16$
  & 0.48 & 0.41 & 0.30 & 0.17 & 0.05\\
\midrule
\multicolumn{16}{@{}l}{\textit{ACC} (\%)}\\
AGC~\cite{AGC}
  & 49.91 & 69.39 & 85.74 & 50.87 & 48.26
  & 51.83 & 46.96 & 50.78 & 48.70 & 47.39
  & 48.00 & 65.74 & 84.35 & 59.65 & 48.26\\
FGC~\cite{FGC}
  & \textbf{100} & 96.09 & 94.35 & 61.30 & 51.30
  & \underline{99.56} & 81.30 & \underline{93.04} & 55.22 & 47.83
  & \underline{99.13} & 92.61 & 91.74 & 55.22 & 45.22\\
SLSA(t)
  & \textbf{100} & 99.13 & 97.39 & 92.61 & 84.78
  & \textbf{99.57} & \underline{96.52} & \textbf{95.65} & 91.30 & \underline{84.35}
  & \textbf{97.75} & 84.67 & 81.48 & 68.70 & 50.09\\
A-CML
  & \textbf{100} & \underline{99.57} & \underline{98.26} & \textbf{96.30} & \textbf{93.48}
  & \textbf{99.57} & \textbf{97.83} & \textbf{95.65} & \textbf{94.78} & \textbf{86.09}
  & \textbf{99.57} & \textbf{97.17} & \textbf{92.61} & \textbf{80.87} & \textbf{71.74}\\
T-CML
  & \textbf{100} & \textbf{100} & \textbf{98.70} & \underline{95.22} & \underline{91.74}
  & \textbf{99.57} & \textbf{97.83} & \textbf{95.65} & \underline{92.39} & \textbf{86.09}
  & \textbf{99.57} & \underline{96.52} & \underline{92.17} & \textbf{80.87} & \underline{70.87}\\
\midrule
\multicolumn{16}{@{}l}{\textit{NMI} (\%)}\\
AGC~\cite{AGC}
  & 19.87 & 27.83 & 56.21 & 9.40 & 7.08
  & 16.42 & 5.19 & 7.91 & 3.82 & 6.06
  & 9.86 & 29.20 & 53.03 & 15.22 & 6.78\\
FGC~\cite{FGC}
  & \textbf{100} & 84.77 & 77.14 & 19.07 & 20.53
  & \textbf{97.75} & 66.57 & 75.30 & 11.49 & 11.30
  & \underline{95.74} & 73.24 & \textbf{71.20} & 10.82 & 7.90\\
SLSA(t)
  & \textbf{100} & 95.64 & 87.45 & 70.99 & 44.84
  & \textbf{97.75} & 84.67 & 81.48 & 68.70 & 50.09
  & 91.97 & \underline{78.18} & 65.18 & 38.08 & 22.75\\
A-CML
  & \textbf{100} & \underline{97.75} & \underline{91.69} & \textbf{84.30} & \textbf{76.29}
  & \textbf{97.75} & \underline{90.51} & \textbf{82.68} & \textbf{79.15} & \underline{55.09}
  & \textbf{97.75} & \textbf{87.40} & \underline{70.89} & \textbf{45.31} & \textbf{28.27}\\
T-CML
  & \textbf{100} & \textbf{100} & \textbf{93.83} & \underline{79.31} & \underline{70.25}
  & \textbf{97.75} & \textbf{91.52} & \underline{82.49} & \underline{71.52} & \textbf{55.81}
  & \textbf{97.75} & \underline{85.31} & 70.56 & \underline{44.29} & \underline{25.90}\\
\midrule
\multicolumn{16}{@{}l}{\textit{F1} (\%)}\\
AGC~\cite{AGC}
  & 42.71 & 68.05 & 85.75 & 48.81 & 46.97
  & 50.99 & 41.14 & 41.91 & 38.21 & 45.96
  & 46.21 & 52.41 & 82.15 & 45.29 & 45.19\\
FGC~\cite{FGC}
  & \textbf{100} & 95.58 & 94.09 & 56.12 & 40.92
  & \textbf{99.62} & 71.74 & 92.41 & 40.53 & 40.04
  & \underline{98.97} & 92.13 & 90.81 & 41.16 & 38.25\\
SLSA(t)
  & \textbf{100} & 99.08 & 97.34 & 92.34 & 80.18
  & \textbf{99.62} & 96.24 & 95.30 & 90.81 & 83.20
  & 98.05 & 94.11 & 89.64 & 73.84 & 66.93\\
A-CML
  & \textbf{100} & \underline{99.63} & \underline{98.38} & \textbf{96.35} & \textbf{94.01}
  & \textbf{99.62} & \textbf{97.56} & \underline{95.41} & \textbf{94.34} & \underline{84.84}
  & \textbf{99.63} & \textbf{96.98} & \textbf{92.25} & \underline{79.08} & \textbf{70.46}\\
T-CML
  & \textbf{100} & \textbf{100} & \textbf{98.88} & \underline{94.87} & \underline{91.76}
  & \textbf{99.62} & \underline{97.38} & \textbf{95.56} & \underline{92.23} & \textbf{85.16}
  & \textbf{99.63} & \underline{96.27} & \underline{92.00} & \textbf{79.71} & \underline{69.72}\\
\bottomrule
\end{tabular}

\end{table}

The structural noise $\delta$ is a bottleneck in attributed graph clustering. For instance, as $\delta$ increases, the performance of FGC is collapses drastically. 
AGC is affected by noise in both $A$ and $X$, as well as by the sparsity of $A$. As shown in Table~\ref{tab:simulate}, the baseline methods are sensitive to errors in the graph structure. 
SLSA(t) also exhibits a continuous and moderate decline as the graph topology degrades, though it is better than FGC. 
A-CML and T-CML maintain remarkable and consistent resilience. Even under extreme structural perturbations (e.g., $\delta=0.9$), both A-CML and T-CML sustain highly competitive results, maintaining a significant performance margin over the baselines.

\begin{table}[t]
\caption{Clustering results under the unbalanced noise scenarios, where
$A(\cdot)$ and $X(\cdot)$ denote the noise levels injected into the graph and the
attributes, respectively, and $\Delta h$ is reported for each scenario. The best and
the second-best results in each column are highlighted in bold and underlined,
respectively.}
\label{tab:unbalanced}
\centering
\fontsize{9}{10}\selectfont
\setlength{\tabcolsep}{6pt}
\begin{tabular}{@{}l*{12}{c}@{}}
\toprule
\multirow{2}{*}{Method}
  & \multicolumn{3}{c}{\shortstack{$A(0.1)$, $X(0.9)$\\$\Delta h{=}0.633$}}
  & \multicolumn{3}{c}{\shortstack{$A(0.3)$, $X(0.9)$\\$\Delta h{=}0.560$}}
  & \multicolumn{3}{c}{\shortstack{$A(0.9)$, $X(0.1)$\\$\Delta h{=}-0.443$}}
  & \multicolumn{3}{c}{\shortstack{$A(0.9)$, $X(0.3)$\\$\Delta h{=}-0.347$}}\\
\cmidrule(lr){2-4}\cmidrule(lr){5-7}\cmidrule(lr){8-10}\cmidrule(lr){11-13}
& ACC & NMI & F1 & ACC & NMI & F1 & ACC & NMI & F1 & ACC & NMI & F1\\
\midrule
AGC~\cite{AGC}
  & 69.74 & 34.87 & 53.89
  & 66.09 & 22.35 & 64.55
  & 53.30 & 10.87 & 52.66
  & 48.26 & 7.08  & 46.97\\
FGC~\cite{FGC}
  & \underline{98.26} & \underline{92.06} & \underline{97.94}
  & 91.30 & 71.88 & 90.39
  & 75.22 & 36.25 & 74.18
  & 51.30 & 20.53 & 40.92\\
SLSA(t)
  & 96.96 & 86.31 & 96.91
  & \underline{93.04} & \underline{75.99} & \underline{92.87}
  & 90.87 & 66.40 & 90.42
  & 84.78 & 52.90 & 84.74\\
A-CML
  & 92.17 & 72.05 & 92.34
  & 85.22 & 56.12 & 83.78
  & \textbf{95.87} & \textbf{84.13} & \textbf{95.71}
  & \textbf{93.48} & \textbf{76.29} & \textbf{94.01}\\
T-CML
  & \textbf{99.13} & \textbf{95.93} & \textbf{99.08}
  & \textbf{95.65} & \textbf{83.51} & \textbf{95.42}
  & \underline{91.52} & \underline{74.39} & \underline{92.87}
  & \underline{91.74} & \underline{70.25} & \underline{91.76}\\
\bottomrule
\end{tabular}
\end{table}

Increasing $\epsilon$ may also hurt performance. However, its negative impact is generally less severe than the structural noise $\delta$. Table~\ref{tab:simulate} shows the results for $\epsilon$ varying from 0.3 to 0.7. 
In this experiment, standard methods rely heavily on accurate graph topology when handling feature distortions.
For instance, FGC achieves competitive results under mild structural noise but loses its effectiveness when $\delta$ exceeds $0.5$.
Both A-CML and T-CML maintain high accuracy ($>90\%$) under high structural noise ($\delta=0.7$), even when attribute noise increases to 0.5.

\begin{table}[ht]
\caption{Clustering results on five real-world datasets. The
\textbf{best}, the \underline{second-best}, and the \uwave{third-best} results in
each column are highlighted.}
\label{tab:realworld}
\centering
\fontsize{9}{10}\selectfont
\setlength{\tabcolsep}{2.5pt}
\begin{tabular}{@{}l*{15}{c}@{}}
\toprule
\multirow{2}{*}{Method} 
  & \multicolumn{3}{c}{Cora}
  & \multicolumn{3}{c}{Citeseer}
  & \multicolumn{3}{c}{Wiki}
  & \multicolumn{3}{c}{ACM}
  & \multicolumn{3}{c}{DBLP}\\
\cmidrule(lr){2-4}\cmidrule(lr){5-7}\cmidrule(lr){8-10}\cmidrule(lr){11-13}\cmidrule(lr){14-16}
&  ACC & NMI & F1 & ACC & NMI & F1 & ACC & NMI & F1
  & ACC & NMI & F1 & ACC & NMI & F1\\
\midrule
K-means      & 33.97 & 15.33 & 30.93 & 42.64 & 19.91 & 40.16 & 33.65 & 23.30 & 29.90 & 79.31 & 44.87 & 79.49 & 42.62 & 34.08 & 42.07\\
Spectral    & 30.00 & 2.36 & 8.78 & 21.60 & 1.54 & 9.46 & 23.20 & 13.74 & 18.05 & 35.57 & 2.06 & 29.36 & 70.22 & 50.94 & 61.47\\
DeepWalk     & 46.74 & 31.75 & 38.06 & 36.15 & 9.66 & 26.70 & 38.46 & 32.38 & 25.74 & 52.33 & 17.56 & 50.24 & 24.28 & 0.04 & 23.86\\
\midrule
GAE \cite{VGAE}        & 53.25 & 40.69 & 41.97 & 41.26 & 18.34 & 29.13 & 17.33 & 11.93 & 15.35 & 84.52 & 55.38 & 84.65 & 61.21 & 30.80 & 61.41\\
VGAE  \cite{VGAE}      & 55.95 & 38.45 & 41.50 & 44.38 & 22.71 & 31.88 & 28.67 & 30.28 & 20.49 & 84.13 & 53.20 & 84.17 & 58.59 & 26.92 & 58.69\\
MGAE   \cite{MGAE}     & 63.43 & 45.57 & 38.01 & 63.56 & 39.75 & 39.49 & 50.14 & 47.97 & 39.20 & 88.12 & 62.16 & 87.01 & 60.25 & 29.84 & 59.68\\
AGC  \cite{AGC}       & 68.92 & 53.68 & 65.61 & 67.00 & 41.13 & 62.48 & 47.65 & 45.28 & 40.36 & 80.05 & 49.68 & 80.11 & 69.58 & 54.45 & 68.76\\
DAEGC \cite{DAEGC}      & 70.40 & 52.80 & 68.20 & 67.20 & 39.70 & 63.60 & 42.66 & 37.87 & 29.05 & 89.72 & 65.56 & 89.70 & 62.02 & 32.49 & 61.75\\
GIC \cite{GIC}        & 70.45 & 52.55 & 68.95 & 68.78 & 43.82 & 64.02 & 46.46 & 48.24 & 40.29 & 90.13 & 68.26 & 88.89 & 75.86 & 44.99 & 73.55\\
S$^2$GC  \cite{SSGC}   & 69.60 & 54.71 & 65.83 & 69.11 & 42.87 & 64.65 & 52.67 & 49.62 & 44.31 & 86.87 & 61.16 & 86.86 & 87.09 & 67.50 & 86.05\\
DFCN   \cite{DFCN}     & 71.40 & 54.61 & 64.29 & 69.50 & 43.90 & 64.30 & 41.60 & 40.00 & 36.10 & 90.15 & 67.98 & 90.14 & 76.00 & 43.70 & 75.70\\
FGC   \cite{FGC}      & 72.12 & 55.21 & 62.71 & 68.16 & 43.15 & 63.47 & 50.89 & 43.27 & 34.60 & 86.12 & 57.94 & 86.02 & \uwave{92.11} & \uwave{76.01} & 91.61\\
GCC  \cite{GCC}       & 73.63 & \underline{57.77} & 69.74 & 69.39 & \underline{45.07} & 64.48 & 54.92 & \underline{53.30} & 45.82 & 65.26 & 54.95 & 53.95 & 89.99 & 69.94 & 89.39\\
SAGSC  \cite{SAGSC}     & 72.71 & 56.54 & 69.26 & 67.23 & 41.92 & 62.74 & 52.56 & 49.22 & 43.20 & \textbf{92.84} & \textbf{73.69} & \textbf{92.87} & 91.22 & 72.61 & 90.72\\
DCGC \cite{DCGC}       & 69.20 & 51.95 & 68.93 & \textbf{71.87} & \textbf{46.87} & \textbf{66.65} & \underline{55.34} & 53.00 & \underline{47.37} & \underline{92.07} & \uwave{71.70} & \uwave{92.10} & 85.73 & 59.43 & 85.28\\
\midrule
CML         & \uwave{74.96} & \uwave{58.09} & \uwave{72.09} & \underline{70.09} & 44.69 & 65.26 & \textbf{56.34} & \textbf{53.40} & \textbf{49.93} & 90.25 & 67.59 & 90.48 & \textbf{93.20} & \textbf{77.81} & \textbf{92.83}\\
CML$^2$     & \textbf{76.62} & \textbf{58.46} & \textbf{74.54} & \uwave{69.97} & \uwave{45.04} & \underline{65.54} & 54.51 & 50.20 & 46.60 & 91.54 & 70.54 & 91.63 & \underline{92.43} & \underline{76.04} & \underline{91.98}\\
CML$^3$     & \underline{75.96} & 56.07 & \underline{74.23} & 69.88 & 44.78 & \uwave{65.26} & \uwave{55.05} & \uwave{53.21} & \uwave{46.88} & \underline{92.07} & \underline{72.34} & \underline{92.33} & 92.05 & 76.00 & \uwave{91.84}\\
\bottomrule
\end{tabular}
\end{table}

Table~\ref{tab:unbalanced} further shows the complementary advantages of A-CML and T-CML on the two noise scenarios:
\begin{center}
small-$\delta$ -- big-$\epsilon$ and big-$\delta$ -- small-$\epsilon$,
\end{center}
depending on which modality remains reliable. When the graph topology is relatively clean, but node attributes are heavily corrupted, T-CML consistently achieves the best performance.
In contrast, when the topology is heavily corrupted while node attributes remain relatively reliable, A-CML becomes the strongest performer, attaining the highest ACC, NMI, and F1-score.
Overall, T-CML effectively exploits trustworthy topological information to suppress severe feature noise, while A-CML is able to learn the clustering structure primarily from the cleaner attribute view.

\subsection{Clustering Performance on Real-World Data}

In practical applications, empirical graph structure $A$ is typically easier to collect and less noisy, whereas node attributes $X$ often have higher dimensionality and more complex noise. Moreover, the homophily difference $\Delta h$ in the \textbf{Supplementary Material} shows consistently higher attribute noise than topological noise on all real-world datasets. Under such asymmetry, we focus on T-CML, which uses the cleaner topology to filter severe feature-level distortions.

We compare with 15 well-known baselines: 12 for attributed graphs, 7 of which are deep-learning based; details are in the \textbf{Supplementary Material}. We apply single, two, and three-stage CML (CML, CML$^2$, CML$^3$) to real-world datasets, followed by Spectral Clustering (SC) on the resulting graphs.

Table \ref{tab:realworld} reports the three clustering metrics on 5 different datasets by 15 baseline methods and our methods. The best results are highlighted in bold, the second best are underlined and the third best are indicated with a wavy line. 
Overall, our CML framework achieves the most competitive and stable clustering performance. Our methods rank first on Cora \cite{cora}, Wiki \cite{wiki}, and DBLP across all three metrics, and consistently place in the top three on F1 across every dataset, demonstrating superior precision‑recall balance. On Cora, our method improves F1 by $\approx 5\%$ over the strongest baseline. While DCGC \cite{DCGC} delivers the best results on Citeseer (with our CML and CML$^2$ following closely), it exhibits notable instability—performing well on Citeseer but poorly on other datasets such as Cora and ACM \cite{acm}. In contrast, our CML variants maintain robust top‑tier performance across all benchmarks.

\begin{figure}[t]
\centering
\begin{minipage}{3cm}
    \centering
    \includegraphics[width=\linewidth]{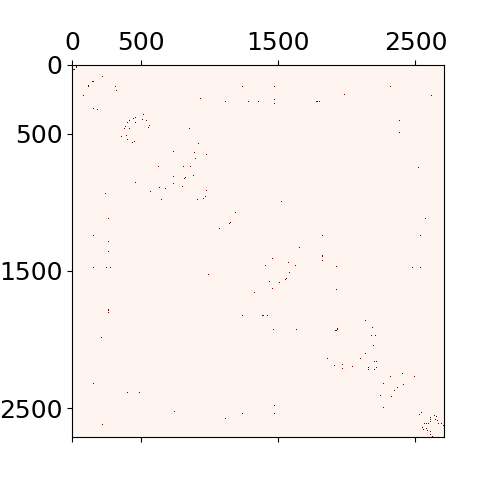}
    \\[-1ex] \scriptsize \(A\)
\end{minipage}
\hspace{-0.2cm}
\begin{minipage}{3cm}
    \centering
    \includegraphics[width=\linewidth]{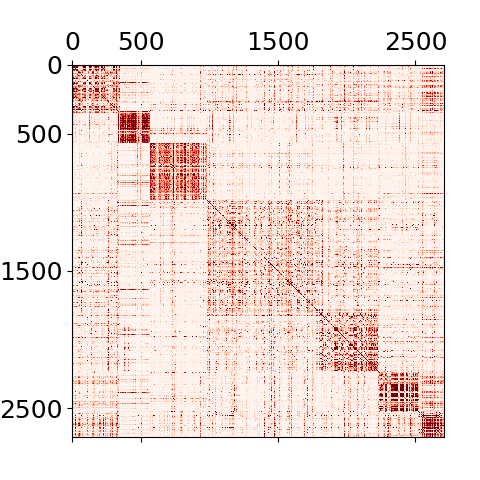}
    \\[-1ex] \scriptsize \(Z^1\)
\end{minipage}
\hspace{-0.2cm}
\begin{minipage}{3cm}
    \centering
    \includegraphics[width=\linewidth]{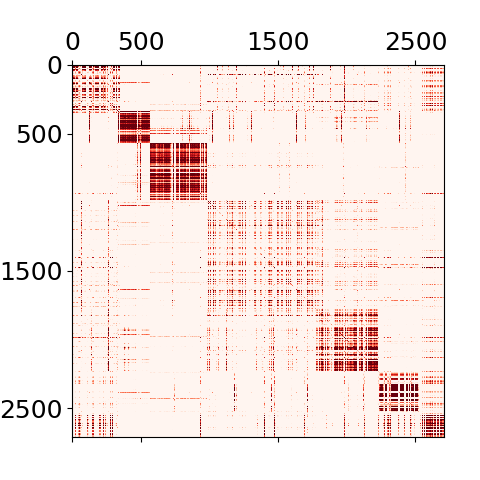}
    \\[-1ex] \scriptsize \(Z^2\)
\end{minipage}
\hspace{-0.2cm}
\begin{minipage}{3cm}
    \centering
    \includegraphics[width=\linewidth]{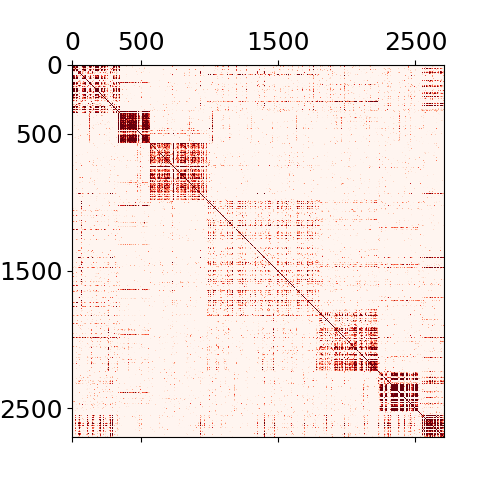}
    \\[-1ex] \scriptsize \(Z^3\)
\end{minipage}
\caption{The leanrned \(Z^1-Z^3\) of multi-stage CML on Cora.}
\label{fig:multi-cora}
\end{figure}

Figure \ref{fig:multi-cora} visualizes $Z^1$, $Z^2$, and $Z^3$ after CML, CML$^2$, and CML$^3$ on Cora. $Z^1$ reveals a coarse community layout; $Z^2$ sharpens the block-diagonal structure by removing cross-cluster noise; $Z^3$ is nearly identical, indicating rapid stabilization within 2–3 stages and diminishing returns from extra iterations. 
\begin{figure}[t!]
\centering
\begin{minipage}{0.32\linewidth}
    \centering
    \includegraphics[width=\linewidth]{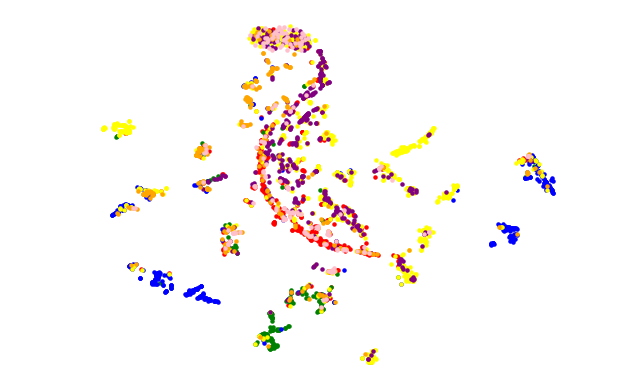}
    \\[-1ex] \scriptsize (a) Cora (raw data)
\end{minipage}
\hfill
\begin{minipage}{0.32\linewidth}
    \centering
    \includegraphics[width=\linewidth]{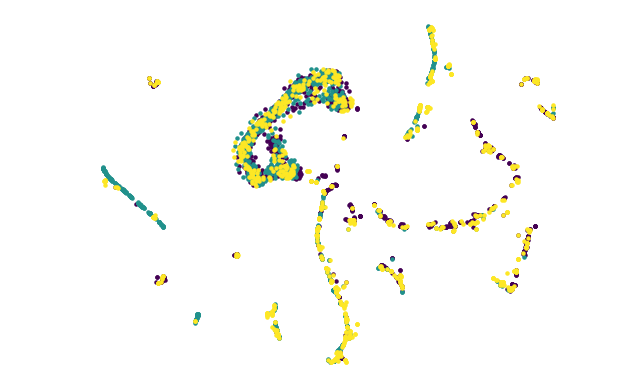}
    \\[-1ex] \scriptsize (b) ACM (raw data)
\end{minipage}
\hfill
\begin{minipage}{0.32\linewidth}
    \centering
    \includegraphics[width=\linewidth]{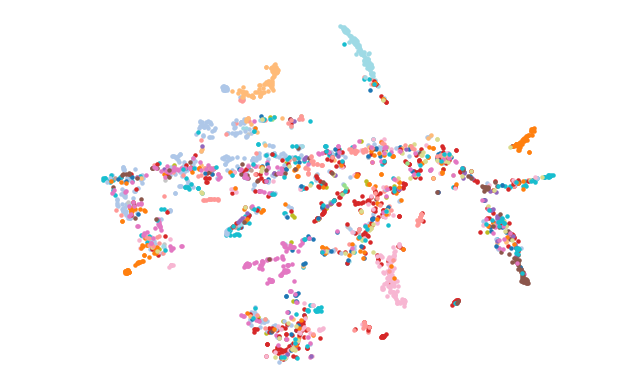}
    \\[-1ex] \scriptsize (c) Wiki (raw data)
\end{minipage}
\begin{minipage}{0.32\linewidth}
    \centering
    \includegraphics[width=\linewidth]{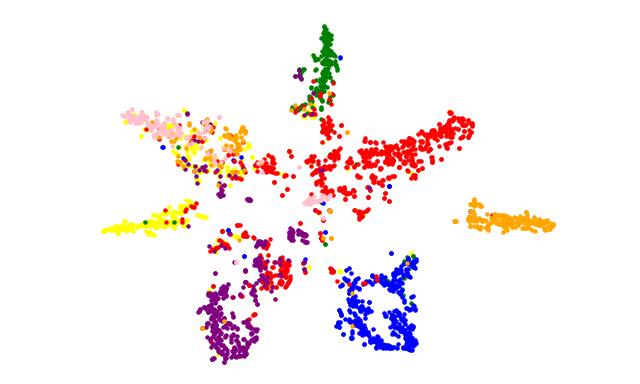}
    \\[-1ex] \scriptsize (d) Cora (CML)
\end{minipage}
\hfill
\begin{minipage}{0.32\linewidth}
    \centering
    \includegraphics[width=\linewidth]{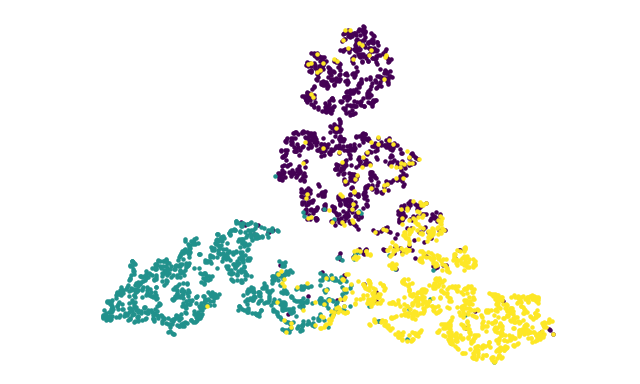}
    \\[-1ex] \scriptsize (e) ACM (CML)
\end{minipage}
\hfill
\begin{minipage}{0.32\linewidth}
    \centering
    \includegraphics[width=\linewidth]{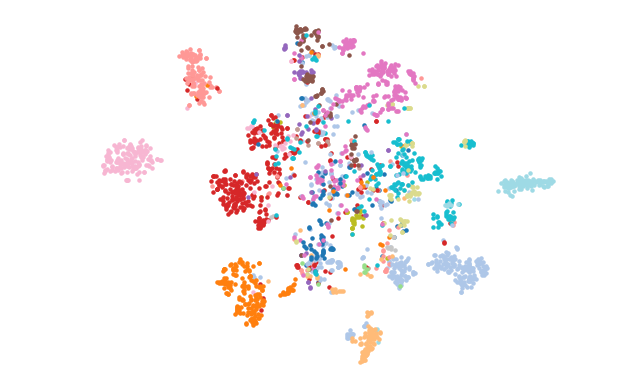}
    \\[-1ex] \scriptsize (f) Wiki (CML)
\end{minipage}
\caption{t-SNE demonstration of the raw and CML node features embedding.}
\label{fig:visual}
\end{figure}

Figure \ref{fig:visual} visualizes the original node attributes in the graph using 2D t-SNE, labeled as \enquote{raw data}, along with their spectral embeddings obtained from the CML graph, labeled as \enquote{CML}, for the Cora, ACM, and Wiki datasets.
The figure shows that the spectral embeddings after CML present much clearer group structure. Interestingly, the embeddings after CML refinement in different classes tend to diverge in different directions, probably due to the low rank and subspace constraint in the CML model. For example, in the Cora dataset with seven categories, the CML embeddings diverge in seven different directions, forming an exquisite heptagon. Similarly, ACM exhibits a triangular pattern due to the divergence of the samples. 

The spectral clustering results of fast-CML on the largest dataset Pubmed are reported in Table \ref{tab:cluster2}.
Since FGC is computationally intractable to run on this dataset, its results are taken from \cite{FGC}. The Fast-CML algorithm can achieve outstanding results with regard to three metrics, with improvement of around $1\%$. With the most competitive performance, the Fast-CML only takes 0.19s for each iteration, while the other methods with similar performance take twice or three times of computation cost than the Fast-CML. For example, GCC takes 3.60$\pm$0.18s and SAGSC takes 4.76$\pm$0.08s for each iteration. The results demonstrate the effectiveness and efficiency of Fast-CML on the large-scale attributed graphs.
\begin{table}[t!]
\caption{The clustering results on Pubmed dataset}
\centering
\fontsize{9}{10}\selectfont
\begin{tabular}{@{}lc|ccc@{}}
\toprule
\multirow{2}{*}{Method}&\multirow{2}{*}{Input}&\multicolumn{3}{c}{Pubmed}\\ 
{}&&ACC$\%$&NMI$\%$&F1$\%$\\
\midrule
K-means & $\textbf{X}$ & 59.51 & 31.26&58.16\\
Spectral & $\textbf{A}$ & 58.96 & 18.30&43.53\\
DeepWalk & $\textbf{A}$ &61.86&16.71&47.06 \\
\midrule
GAE& $\textbf{A,X}$&64.08 &22.97 &49.26\\
VGAE& $\textbf{A,X}$&65.48 &25.09 &50.95\\
MGAE& $\textbf{A,X}$&43.88 &8.16 &41.98\\
AGC& $\textbf{A,X}$ &69.78&31.59&68.72	\\
DAEGC& $\textbf{A,X}$ &67.10&26.60&65.90\\
GIC& $\textbf{A,X}$&64.30&26.02&64.86\\
$\text{S}^2\text{GC}$(2021)& $\textbf{A,X}$&\uwave{70.74}&{32.49}&\uwave{69.97}\\
DFCN& $\textbf{A,X}$&68.89 &31.43 &68.10\\
FGC 
& $\textbf{A,X}$&	{70.01}&31.45&{69.10}\\
GCC &$\textbf{A,X}$ &{68.65} &\underline{33.09} &{68.23}\\
SAGSC & $\textbf{A,X}$& \underline{71.00} & \uwave{32.97} &\underline{70.11}\\
DCGC & $\textbf{A,X}$& 67.05 &28.33 &67.59\\
Fast-CML (Ours)& $\textbf{A,X}$ &\bf{71.94}&\bf{33.31}&\bf{71.02}\\
\bottomrule
\end{tabular}
\label{tab:cluster2}
\end{table}

\subsection{CML for Single-input Clustering}

When only a single input modality (either the adjacency matrix $A$ or the attribute matrix $X$) is available, clustering must rely on methods designed for single-input data. For cross-modal methods requiring both the adjacency matrix and the attribute matrix, two strategies are adopted to address the data deficiency:
\begin{itemize}
    \item If an adjacency matrix $A$ is available only, set $X=A$ by directly using the adjacency matrix as the attribute matrix.
    \item If attribute $X$ is available only, construct an adjacency matrix $A$ from $X$. 
\end{itemize}
For the second strategy, the adjacency matrix can be constructed either from a neighborhood graph of the attribute vectors or from a symmetric nonnegative matrix derived from the self-representation of $X$ in subspace learning.

In this subsection, we show that CML consistently outperforms both single-input methods and conventional cross-modal methods under these substitution strategies.

\subsubsection{CML with Adjacency Matrix Only}

\begin{table*}[t]
\caption{Comparison of CML and 
subspace methods in clustering task.}
\centering
\fontsize{9}{10}\selectfont
\setlength{\tabcolsep}{4pt}
\begin{tabular}{lcc|ccc|ccc|ccc}
\toprule
{Input}& Modal type & Method 
    & ACC  & NMI  & F1    
    & ACC  & NMI  & F1 
    & ACC  & NMI  & F1 \\ \midrule
 & & &\multicolumn{3}{c}{Cora} 
    & \multicolumn{3}{c}{Citeseer} 
    & \multicolumn{3}{c}{ACM}  \\\cmidrule{2-12}
\multirow{4}{*}{$A$} & Single modality & DeepWalk \cite{DeepWalk} 
    &46.74 &31.75 &38.06 &36.15 &9.66 &26.70 
    &52.33 &17.56 &50.24\\ \cmidrule{2-12}
&\multirow{2}{*}{Multi-modality} & FGC   
    & 68.72 & 50.49 & 66.15 & 43.16 & 19.18 
    & 30.98 & 56.26 & 22.97 & 53.78 \\
&\multirow{2}{*}{($X:=A$)} & DCGC   
    & 52.99 & 34.66 & 52.70 & 46.32 & 21.93 
    & 44.99 & 59.93 & 21.51 & 60.47 \\
&   &T-CML & 71.05 & 52.07 & 68.28 & 46.77 
    & 25.67 & 43.67 &  60.56 & 22.88 & 59.86 \\\midrule
 & & &\multicolumn{3}{c}{Iris} 
    &\multicolumn{3}{c}{Yale} 
    &\multicolumn{3}{c}{Digits} \\ 
    \cmidrule{2-12}
  &\multirow{2}{*}{Single modality} &{SSC}   &74.67&59.72&74.66   &25.45&23.39&24.22 &LRT&LRT&LRT\\
\multirow{10}{*}{$X$} & 
    &{SSCE} &89.33&74.66&89.24  &27.88&23.28&28.41  &78.58&72.89 &78.68\\\cmidrule{2-12}
&\multirow{2}{*}{Multi-modality}
    &{FGC} &92.00 &76.01 &92.04  &27.28 &20.57 &26.77  &79.02 &73.56 &78.72\\
&\multirow{2}{*}{(Gaussian $A$)}
    & DCGC &97.33 &91.44 &97.33  &29.09 &24.86 &29.32   &82.58 &73.34 &82.86\\
&   &T-CML  &98.67 &94.88 &98.67  &33.94 &30.15 &34.60             &83.70 &75.04 &84.94 \\\cmidrule{2-12}
&\multirow{2}{*}{Multi-modality}
    &{FGC}&82.67&65.73&82.56  &26.67&20.67&25.13 &77.69&72.80&77.63\\
&\multirow{2}{*}{(Linear $A$)}
    & DCGC &{97.33}&{91.44} &{97.33} &29.70&24.91&29.27   &83.70&73.54&83.54\\
&   &T-CML &{95.33}&{84.98}&{95.33}  &35.15&26.44&34.66            &81.41&73.87&81.09\\ \cmidrule{2-12}
&{Multi-modality} & {SLSA} &91.33 &77.24 &91.31  
    &30.30&23.80&31.31  &79.86&73.41&79.93\\
&{($A$ := SSCE)} & {T-CML}  &95.76&86.57 &95.75       
    &32.12&28.60&32.61   &86.70 &80.35 &86.48\\
\bottomrule
\end{tabular}
\label{tab:subspace1}
\end{table*}

When an adjacency matrix $A$ is available based on structural connections, we use the simple copy $X = A$ as a substitute of deficient attribute matrix.
Table \ref{tab:subspace1} reports the clustering performance of this strategy on three real-world datasets (Cora, Citeseer, and ACM) using only the adjacency matrices. Compared with the multi-modal results in Table \ref{tab:realworld}, all methods experience a noticeable performance drop under this single-modal setting. This disparity underscores that node attributes and adjacency matrices provide significant complementary information for attributed graph clustering. Furthermore, even within this constrained single-modal scenario, our CML consistently outperforms the competing baselines, demonstrating its superior capability to fully uncover hidden structural information.

\subsubsection{CML with Node Attributes Only}

When topology is absent and only multi-dimensional attributes $X$ are available, we construct $A$ from raw $X$ via Gaussian RBF, linear kernel, and SSCE \cite{SSCE} subspace self-representation to capture pairwise geometric similarities.

We compare our CML method with two baseline subspace learning models SSC~\cite{SSC} and SSCE~\cite{SSCE},
and the cross-modal methods FGC \cite{FGC} and DCGC \cite{DCGC} on three benchmark datasets: Yale,
Iris \cite{iris}\footnote{\url{https://scikit-learn.org/stable/modules/generated/sklearn.datasets.load_iris.html}} and Digits\footnote{\url{https://scikit-learn.org/stable/modules/generated/sklearn.datasets.load_digits.html}}, which contain only features and category information for the samples. The details of datasets and baseline methods are described in the \textbf{Supplementary Material}.


Table~\ref{tab:subspace1} reports the clustering performance of these methods. For the
Gaussian kernel, multiple values of $\sigma$ are evaluated, and the best-performing
result is reported. For SSC and SSCE, different regularization parameters $\lambda$
are considered, and the corresponding best result is reported. Overall, a T-CML variant attains the best ACC, NMI, and F1 on each of the three
datasets, with the winning construction of $A$ depending on the dataset: on Iris the
Gaussian kernel leads DCGC by 1.34\%/3.44\%/1.34\%, on Yale all three variants rank
above every baseline (up to 5.45\%/5.24\%/5.34\% over DCGC), and on Digits the
SSCE-constructed $A$ gives 3.00\%/6.81\%/2.94\%. SSC fails on Digits and is reported
as LRT (long runtime).

\subsection{Improving Graph Clustering Methods by CML}
Generally, graph clustering methods based on cross-modal models like FGC, AGC, and SAGSC \cite{SAGSC} perform clustering attributed graph nodes, depending on the quality of the adjacency matrix $A$ and the attribute matrix $X$. Using the learned graph $Z$ by CML as a replacement adjacency matrix, the clustering accuracy of these methods can be significantly improved. We denote by CML+M the combination of CML and a method M (FGC, AGC, or SAGSC), where $A$ is replaced by $Z^1$; CML$^2$+M uses $Z^2$ instead. Here $Z^1$ and $Z^2$ are refined graphs from single- and two-stage CML, with $X$ unchanged. Table \ref{tab:CML+*} reports results on Cora, Citeseer, Wiki, and ACM.

\begin{table*}[t]
\centering
\caption{Clustering performance of FGC, AGC, and SAGSC using original or CML-improved graphs.}
\label{tab:rcs_improvements_merged}
\fontsize{9}{10}\selectfont
\setlength{\tabcolsep}{4pt}
\begin{tabular}{l|ccc|ccc|ccc|ccc@{}}
\toprule
\multirow{2}{*}{Method} 
& \multicolumn{3}{c|}{Cora} 
& \multicolumn{3}{c|}{Citeseer} 
& \multicolumn{3}{c|}{Wiki} 
& \multicolumn{3}{c}{ACM} \\
&  ACC  & NMI  & F1  & ACC  & NMI  & F1  & ACC  & NMI  & F1  & ACC  & NMI  & F1  \\
\midrule  
\bfseries FGC & 72.12 & 55.21 & 62.71 & 68.16 & 43.15 & 63.47 & 50.89 & 43.27 & 34.60 & 86.12 & 57.94 & 86.02 \\
CML+FGC  &  {74.41} & \bf{58.65} &  {72.19} & 65.80 & 40.25 & 61.47 & \bf{55.63} &  {53.69} & \bf{46.58} &  {91.14} &  {69.70} &  {91.20} \\
CML$^2$+FGC  & \bf{76.62} &  {58.35} & \bf{74.88} & \bf{69.70} & \bf{44.16} & \bf{65.28} &  {54.97} & \bf{54.74} &  {46.08} & \bf{92.36} & \bf{72.56} & \bf{92.40} \\

\midrule  
\bfseries AGC  & 68.92 & 53.68 & 65.61 & 67.00 & 41.13 & 62.48 & 47.65 & 45.28 & 40.36 & 80.05 & 49.68 & 80.11 \\
CML+AGC &  {73.67} &  {57.35} &  {72.05} & 66.31 & 41.11 & 62.33 &  {53.44} &  {53.28} & 45.17 & 91.03 & 69.27 & 91.12 \\
CML$^2$+AGC & \bf{75.37} & \bf{58.92} & \bf{73.77} &  \bf{68.98} &  \bf{43.40} &  \bf{64.68} & \bf{54.15} & \bf{54.31} & \bf{46.99} & \bf{92.23} & \bf{72.26} & \bf{92.26} \\

\midrule  
\bfseries SAGSC  & 72.71 & 56.54 & 69.26 &  \bf{67.23} & \bf{41.92} &  \bf{62.74} & 52.56 & 49.22 & 43.20 & 92.84 & 73.69 & 92.87 \\

CML+SAGSC  &  {74.74} & \bf{59.48} &  {72.64} & 65.69 & 40.56 & 62.07 &  {54.40} &  {51.70} &  {44.40} & \bf{93.28} & \bf{75.13} &  {93.21} \\
CML$^2$+SAGSC  & \bf{75.22} &  {58.87} & \bf{73.66} & 65.06 & 39.39 & 60.65 & \bf{54.62} & \bf{52.04} & \bf{44.71} &  {93.25} &  {75.07} & \bf{93.26} \\
\bottomrule
\end{tabular}
\label{tab:CML+*}
\end{table*}

For FGC, $Z^2$ generally yields the largest gains. On Cora, CML$^2$+FGC improves ACC/NMI/F1 by 4.5\%/3.14\%/12.17\% over FGC; on Wiki and ACM, CML-refined-graphs also raise ACC/NMI/F1 above 5\%. Thus a cleaner graph structure translates into higher clustering accuracy. AGC also benefits: CML$^2$+AGC improves Cora by 6.45\%/5.24\%/8.16\%, Wiki by 6.5\%/9.03\%/6.63\%, and ACM by 12.18\%/22.58\%/12.15\% in ACC/NMI/F1. SAGSC gains are smaller and dataset-dependent: CML improve Cora, Wiki, and ACM modestly, but are lower than vanilla SAGSC on Citeseer. Overall, CML improves existing graph clustering methods in most cases, while the gain depends on the base method and dataset.

\subsection{Parameter Sensitivity}

\begin{table}[t]
\centering
\caption{Parameter Sensitivity of $\alpha$ in A-CML and T-CML.}
\fontsize{9}{10}\selectfont
\setlength{\tabcolsep}{3pt}
\begin{tabular}{@{}ccccccccccccc@{}}
\toprule
Data & Methods & Parameter & ACC\% & NMI\% & F1\% && Data & Methods & Parameter & ACC\% & NMI\% & F1\% \\
\midrule
\multirow{5}{*}{A(0.1)} & \multirow{3}{*}{A-CML} & $\alpha=0.1$ & 90.65 & 70.92 & 91.73 && \multirow{5}{*}{A(0.9)} & \multirow{3}{*}{\bf{A-CML}} & $\alpha=0.1$ & 39.57 & 1.73 & 37.64 \\
\multirow{5}{*}{X(0.9)}  &  & $\boldsymbol{\alpha=1}$ & \bf{92.17} & \bf{72.05} & \bf{92.34}   &&\multirow{5}{*}{X(0.1)} & & $\alpha=1$ & 81.74 & 46.27 & 80.90 \\
 &  & $\alpha=10$ & 84.35 & 55.39 & 83.89 &&  &  & $\boldsymbol{\alpha=10}$ & \bf{93.04} & \bf{73.77} & \bf{92.70} \\
\cmidrule{2-6} \cmidrule{9-13}
& \multirow{3}{*}{\bf{T-CML}} & $\alpha=0.1$ & 98.91 & 95.00 & 98.67 &&  & \multirow{3}{*}{T-CML} & $\alpha=0.1$ & 46.52 & 6.02 & 40.99 \\
 &  & $\boldsymbol{\alpha=1}$ & \bf{99.13} & \bf{95.74} & \bf{98.96} &&  &  & $\boldsymbol{\alpha=1}$ & \bf{91.52} & \bf{68.96} & 91.08 \\
 &  & $\alpha=10$ & 96.52 & 84.45 & 96.30 &&  &  & $\alpha=10$ & 91.30 & 68.67 & \bf{91.09} \\
\bottomrule
\end{tabular}
\label{tab:CMLpara}
\end{table}

\begin{figure*}[t]
    \centering
    \begin{minipage}{5cm}
        \centering
        \includegraphics[width=5cm]{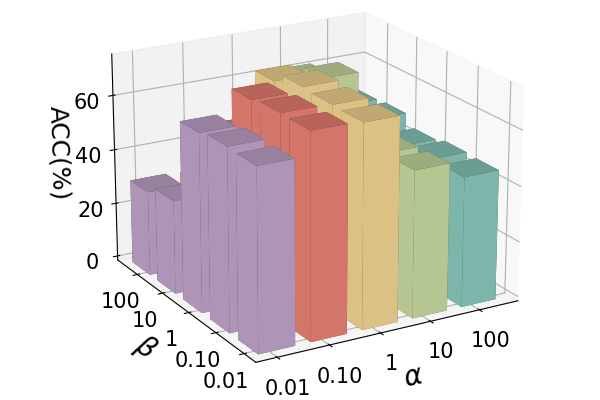}
    \end{minipage}
    \hspace{-0.5cm}
    \begin{minipage}{5cm}
        \centering
        \includegraphics[width=5cm]{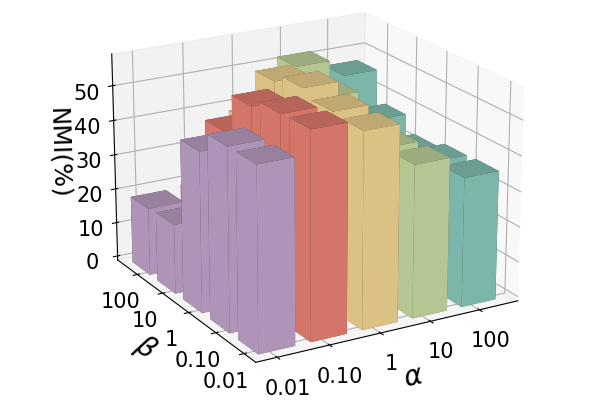}
    \end{minipage}
    \hspace{-0.5cm}
    \begin{minipage}{5cm}
        \centering
        \includegraphics[width=5cm]{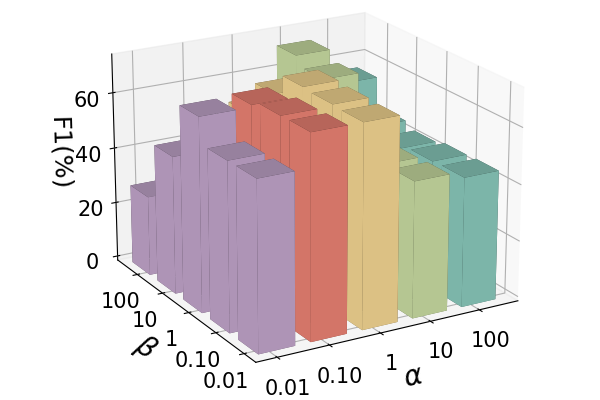}
    \end{minipage}
    \caption{The effect of parameters $\alpha$ and $\beta$ on the results in the experiments on the Cora dataset.}
    \label{fig:parameter a,b}
\end{figure*}

In both A-CML and T-CML, $\alpha$ explicitly controls the weight of the attribute domain, while other parameters controlling the topology domain remain fixed at 1.
Table \ref{tab:CMLpara} shows that the parameter trade-off aligns perfectly with the relative noise levels of the two modalities. When the graph topology is clean but attributes are noisy ($A(0.1), X(0.9)$), performance inherently favors T-CML  
with $\alpha = 1$. This setting ensures that heavily contaminated features do not dominate 
pristine structural information.  
In contrast, when the topology is heavily corrupted but the attributes are clean ($A(0.9), X(0.1)$), the preference switches to A-CML, which peaks at a higher  
value ($\alpha = 10$). By increasing $\alpha$, the framework effectively assigns a much higher priority to the reliable attribute domain, successfully mitigating structural noise from the topology domain.
This distinct trend offers a practical selection criterion for real-world applications, suggesting that T-CML is preferred with moderate $\alpha$ (e.g. $\alpha = 1$) when graph structures are highly reliable, while 
A-CML with a higher $\alpha$ should be selected to suppress heavy topological noise.

\begin{figure*}[t]
    \centering
    \begin{minipage}{5cm}
        \centering
        \includegraphics[width= 5cm]{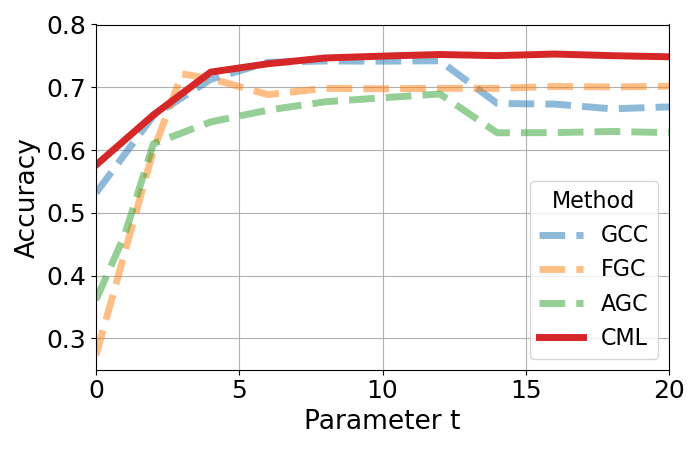}
    \end{minipage}
    \hspace{-0.2cm}
    \begin{minipage}{ 5cm}
        \centering
        \includegraphics[width= 5cm]{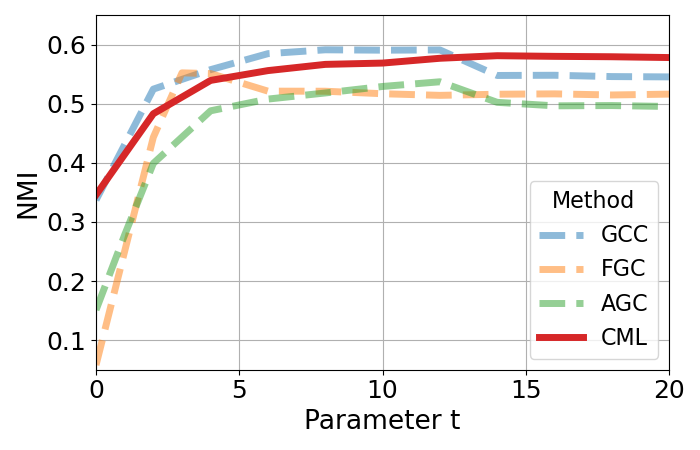}
    \end{minipage}
    \hspace{-0.2cm}
    \begin{minipage}{ 5cm}
        \centering
        \includegraphics[width= 5cm]{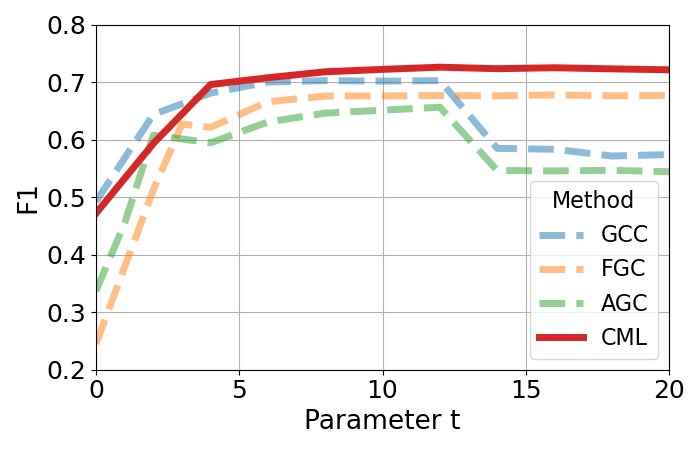}
    \end{minipage}
    \caption{Sensitivity of clustering methods to the attribute smoothing parameter $t$ during preprocessing.}
    \label{fig:parameter t}
\end{figure*}

Figure \ref{fig:parameter a,b} further shows the dependence on parameter $\beta$. An evident phenomenon is that performance tends to degrade when there is a large magnitude difference between $\alpha$ and $\beta$, or when both are  identically small or large. The best performance is achieved when $\beta$ is of similar magnitude to a fixed $\alpha$, with both parameters in the range [0.1, 1]. This is because the constant coefficient of $\|Z - A\|_F$ and the rescaling of $A$ and $X$ require $\alpha$ and $\beta$ to be of comparable magnitude, ideally around 1.  

The parameter $t$ low-pass graph filter mentioned in \textit{Section 6.1.2} may also affect the performance. Figure \ref{fig:parameter t} shows that our CML method performs more stably as $t$ increases, compared with FGC, GCC, and AGC. 

\section{Conclusion}
This study introduces a novel Cross-Modality Learning (CML) framework for recovering latent clustering structures from heavily corrupted attributed graphs. By exploiting the complementary information between graph topology and node attributes, the CML framework performs bidirectional cross-modality recovery, enabling the relatively clean modality to restore the clustering structure of the severely corrupted one. Within this framework, we develop two complementary models, A-CML and T-CML, using attribute and topological information to guide structure recovery, respectively.

Extensive experiments on both synthetic and real-world datasets demonstrate the effectiveness and robustness of the proposed framework. In particular, results under highly asymmetric and extreme noise settings confirm that A-CML and T-CML can adaptively exploit the cleaner modality to recover the corrupted one, leading to substantial performance gains over existing methods. Moreover, comprehensive evaluations on attributed graph clustering and high-dimensional subspace clustering benchmarks show that the CML consistently outperforms state-of-the-art competitors. Overall, these findings highlight the potential of cross-modality recovery for learning reliable clustering structures from noisy data and suggest a promising direction for integrating structure recovery mechanisms with graph neural networks and other graph representation learning models.

\section*{Acknowledgments}

This work was funded by National Natural Science Foundation of China project under Grant No. 92470106 and National Key Research and Development Program of China (Grant No. 2025YFE0113400).



%
\bibliographystyle{elsarticle-num}
\bibliography{refs}

\end{document}